\documentclass[10pt,conference]{IEEEtran}
\IEEEoverridecommandlockouts
\usepackage{cite}
\usepackage{amsmath,amssymb,amsfonts} 
\usepackage{algorithmic}
\usepackage[caption=false,font=footnotesize]{subfig}
\usepackage{textcomp}
\usepackage{xcolor}
\usepackage{longtable}
\usepackage{rotating}
\usepackage{url}
\usepackage{hyperref}
\usepackage{multirow}           
\usepackage{booktabs}           
\usepackage{graphicx}           
\usepackage{array}

\newcolumntype{C}[1]{>{\centering\arraybackslash}p{#1}}

\makeatletter
\renewcommand{\@seccntformat}[1]{%
  \csname the#1\endcsname\quad}  
\makeatother

\def\BibTeX{{\rm B\kern-.05em{\sc i\kern-.025em b}\kern-.08em
    T\kern-.1667em\lower.7ex\hbox{E}\kern-.125emX}}
\usepackage{placeins}
\begin{document}

\title{Transcriptome-Conditioned Personalized De Novo Drug Generation for AML Using Metaheuristic Assembly and Target-Driven Filtering}

\author{
\IEEEauthorblockN{
Abdullah G. Elafifi\textsuperscript{1}\,%
\href{https://orcid.org/0009-0007-0424-1425}{\includegraphics[width=10pt]{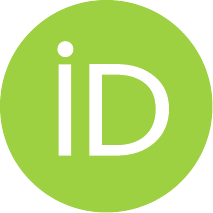}}, 
Basma Mamdouh\textsuperscript{2}, 
Mariam Hanafy\textsuperscript{3}, 
Muhammed Alaa Eldin\textsuperscript{4}, \\
Yosef Khaled\textsuperscript{5}, 
Nesma Mohamed El-Gelany\textsuperscript{6}, 
Tarek H.M. Abou-El-Enien\textsuperscript{7}\,%
\href{https://orcid.org/0000-0001-5345-8873}{\includegraphics[width=10pt]{orcid.pdf}}, 
}

\vspace{0.3cm}

\IEEEauthorblockA{
\textsuperscript{1}20221238@stud.fci-cu.edu.eg, 
\textsuperscript{2}11410120221036@stud.cu.edu.eg, 
\textsuperscript{3}11410120221146@stud.cu.edu.eg, \\
\textsuperscript{4}20211083@stud.fci-cu.edu.eg, 
\textsuperscript{5}11410120221244@stud.cu.edu.eg, \\
\textsuperscript{6}n.mohamed@fci-cu.edu.eg, 
\textsuperscript{7}t.hanafy@fci-cu.edu.eg
}

\vspace{0.2cm}

\IEEEauthorblockA{
\textsuperscript{1--7} Department of Operations Research and Decision Support,\\
Faculty of Computers and Artificial Intelligence, Cairo University, Giza, Egypt
}
}

\maketitle
\section{\textbf{Abstract}}

Acute Myeloid Leukemia (AML) remains a clinical challenge due to its extreme molecular heterogeneity and high relapse rates. While precision medicine has introduced mutation-specific therapies, many patients still lack effective, personalized options. This paper presents a novel, end-to-end computational framework that bridges the gap between patient-specific transcriptomics and de novo drug discovery.
By analyzing bulk RNA sequencing data from the TCGA-LAML cohort, the study utilized Weighted Gene Co-expression Network Analysis (WGCNA) to prioritize 20 high-value biomarkers, including metabolic transporters like HK3 and immune-modulatory receptors such as SIGLEC9. The physical structures of these targets were modeled using AlphaFold3, and druggable hotspots were quantitatively mapped via the DOGSiteScorer engine. Then developed a novel, a reaction-first evolutionary metaheuristic algorithm as well as multi-objective optimization programming that assembles novel ligands from fragment libraries, guided by spatial alignment to these identified hotspots. 
The generative model produced structurally unique chemical entities with a strong bias toward drug-like space, as evidenced by QED scores peaking between 0.5 and 0.7. Validation through ADMET profiling and SwissDock molecular docking identified high-confidence candidates, such as Ligand L1, which achieved a binding free energy of -6.571 kcal/mol against the A08A96 biomarker. These results demonstrate that integrating systems biology with metaheuristic molecular assembly can produce pharmacologically viable, patient-tailored leads, offering a scalable blueprint for precision oncology in AML and beyond.
\\
\begin{IEEEkeywords}
Precision oncology, Acute Myeloid Leukemia (AML), Transcriptomics, De novo drug generation, Computational drug design, Metaheuristic Assembly, Ligand generation.
\end{IEEEkeywords}

\section{\textbf{Introduction}}
Acute Myeloid Leukemia (AML) is an aggressive hematological malignancy characterized by uncontrolled proliferation of myeloblasts in the bone marrow, leading to rapid disease progression if untreated. The accumulation of leukemic cells disrupts normal hematopoiesis, resulting in anemia, infections, bleeding disorders, and, in advanced cases, infiltration of peripheral blood and extramedullary tissues [1]. AML arises from genetic alterations in hematopoietic progenitor cells, although the precise mechanisms driving these mutations remain incompletely understood. A combination of environmental exposures, prior therapeutic interventions, and inherited predispositions has been associated with increased disease risk [1].
AML treatment strategies include intensive chemotherapy, radiotherapy, hematopoietic stem cell transplantation, and targeted therapies, with treatment selection guided by disease subtype, molecular characteristics, and patient fitness. Standard therapy is typically divided into induction and post-remission phases. Induction therapy aims to achieve complete remission by eliminating leukemic blasts, while post-remission therapy targets residual disease to prevent relapse [1]. Clinically, AML is diagnosed when myeloblasts constitute at least 20
Despite therapeutic advances, AML outcomes remain poor. Global incidence increases with age, with an estimated 3–5 cases per 100,000 individuals annually, accounting for nearly one-third of adult leukemia cases. Epidemiological data from the Surveillance, Epidemiology, and End Results (SEER) program project approximately 22,010 new AML cases and 11,090 AML-related deaths in the United States in 2025, with a 5-year relative survival rate of only 32.9
Moreover, the implementation of intensive chemotherapy is highly resource dependent. High treatment costs, prolonged hospitalization, and the need for specialized supportive care limit accessibility in low- and middle-income settings. As a result, therapeutic strategies must be adapted to local healthcare capacities to ensure safe and equitable AML management [3]. These limitations highlight the urgent need for more effective, accessible, and individualized treatment approaches.
Precision medicine has emerged as a transformative paradigm in oncology, shifting treatment strategies from uniform protocols toward patient-specific approaches informed by molecular tumor characteristics [5]. Advances in high-throughput sequencing technologies have enabled comprehensive profiling of genetic and transcriptomic alterations, while patient-derived experimental models support functional drug testing and validation. Together, these developments have expanded the therapeutic landscape to include targeted agents, antibody–drug conjugates, and emerging modalities such as gene- and mRNA-based therapies [5]. Effective implementation of precision oncology requires scalable computational frameworks capable of integrating molecular data with predictive modeling and robust validation strategies.

In AML, transcriptomic profiling has enabled detailed characterization of patient-specific gene expression patterns, facilitating the identification of dysregulated genes and pathways that may serve as therapeutic targets. Incorporating transcriptomic information into computational drug design pipelines provides an opportunity to guide ligand generation toward molecular features specific to individual patients, potentially improving efficacy while minimizing off-target effects.
Structure-based de novo drug design aims to generate novel ligands directly from the three-dimensional structure of a target protein, independent of prior ligand knowledge [6]. However, the vastness of chemical space and the sparsity of bioactive regions present significant challenges [6]. Computational strategies address these challenges by constraining the search space through fragment-based assembly and structure-guided exploration of chemically compatible substructures [6].
Evolutionary and genetic algorithms have been widely applied to de novo drug design due to their ability to efficiently explore large, complex, and discontinuous search spaces [6]. These population-based metaheuristic methods iteratively generate candidate molecules, evaluate their fitness, and evolve improved solutions through stochastic processes such as selection, crossover, and mutation [7]. Importantly, evolutionary algorithms enable simultaneous optimization of multiple objectives, including biological activity, drug-likeness, ADMET properties, and synthetic feasibility, and can identify Pareto-optimal solutions when objectives conflict [6 ] [7].

The increasing cost, duration, and high attrition rates of traditional drug discovery pipelines have driven the adoption of Computer-Aided Drug Discovery (CADD) methodologies [8]. Advances in structural biology have made high-quality macromolecular structures increasingly available, establishing structure-based approaches as central components of modern drug design [8]. While ligand-based drug design remains valuable when structural data are unavailable, its effectiveness depends on existing ligand diversity and is limited by alignment and conformational sampling challenges [9]. When reliable target structures are accessible, structure-based strategies provide a more direct and informative framework for rational ligand design [9].
By integrating patient-specific transcriptomic profiles with structure-based de novo drug design, evolutionary optimization, and target-driven filtering, this work proposes a personalized computational framework for AML therapeutics. This approach combines molecular specificity, structural insight, and computational efficiency to address key limitations of conventional treatment strategies and advance precision medicine for AML.
\\

\section{\textbf{Literature Review}}
AML is molecularly and clinically heterogeneous, which has motivated precision approaches to therapy. Recurrent driver alterations FLT3-ITD, NPM1, DNMT3A define subgroups with distinct biology, prognosis, and drug-sensitivity profiles; moreover, clonal evolution and treatment-induced resistance remain major challenges for long-term disease control [10]. Standard-of-care evolution in AML now includes small-molecule targeted agents such as the most popular one FLT3 inhibitors, epigenetic modulators, and BCL-2 inhibition in combination regimens, but responses are variable and relapse is common, emphasizing the need for personalized discovery strategies [11]. 
Using gene-expression signatures to connect diseases, genes, and compounds is a well-established paradigm that underpins transcriptome-conditioned drug discovery. The original Connectivity Map demonstrated that perturbation expression signatures can reveal mechanistic links between small molecules and disease states, enabling signature-reversal repurposing strategies [12]. The Library of Integrated Network-Based Cellular Signatures program dramatically expanded this idea via the L1000 platform, producing large compendia of compound and perturbation-induced signatures useful for connectivity scoring and signature inversion workflows. Tools and derivative resources such as CREEDS and L1000FWD provide precomputed signatures and visualization/search interfaces that are commonly used for expression-driven repurposing and hypothesis generation [12], [13]. These methods form a natural foundation for conditioning molecule design on patient transcriptomes but generally stop short of generating new chemotypes tailored to individual signatures. 
While having these difficulties with generating new chemotype, the translating transcriptomic perturbation into actionable targets typically requires integration with orthogonal resources. Systematic genetic screens RNAi/CRISPR and the Cancer Dependency Map have become critical for identifying essential genes and context-specific vulnerabilities, while knowledgebases such as OncoKB provide clinical annotations linking genomic events to therapeutic strategies [14]. Network and pathway approaches  are widely used to prioritize target candidates from patient-level omics. These resources make it feasible to connect patient signatures with mechanistic targets for downstream filtering or scoring of designed molecules. 
The de novo molecular generation has evolved rapidly, spanning SMILES-based language models, variational autoencoders (VAEs), generative adversarial networks (GANs), and graph-based deep generative models [15]. Work by Gómez-Bombarelli et al. established continuous latent-space VAEs for molecule optimization; subsequent graph-focused architectures such as the Junction Tree VAE advanced generation of chemically valid molecular graphs by assembling scaffolds and substructures [16]. Implicit graph generative models (MolGAN) and many later graph-neural approaches improved chemical validity and property control for generated compounds [17]. Large chemical language models for SMILES/SELFIES transformers and conditional generation techniques have further broadened the design toolkit. While powerful, most deep generative methods are trained on a population-level chemistry and are optimized toward physicochemical objectives rather than patient-specific biology. 
Fragment-based drug discovery (FBDD) is an established strategy in medicinal chemistry that screens small, low-molecular-weight fragments and grows or links them into potent leads. Computational fragment enumeration and recombination often using retrosynthetic fragmentation rules such as BRICS are used to create chemically sensible fragment libraries and enable scaffold assembly approaches in silico [18], [19]. Fragment-aware generative and combinatorial assembly methods allow exploration of chemical space with more synthetically plausible building blocks and are naturally compatible with metaheuristic recombination strategies such as linking, and crossover of fragment sets
The evolutionary and metaheuristic algorithms have a long history in computational molecular design and remain competitive with deep generative models for certain tasks. Recent work has adapted graph-based genetic algorithms, multi-objective evolutionary optimization and hybrid approaches that combine genetic search with learned models or Monte Carlo tree search [20]. Evolutionary/metaheuristic methods are particularly appealing when the objective function is complex because they can directly incorporate docking, synthetic-accessibility filters, and domain-specific crossover/mutation rules for fragments [21]. These characteristics make metaheuristics a natural fit for fragment-assembly pipelines conditioned on biological objectives. 
Post-generation filtering and ranking are essential to practical de novo pipelines. Traditional structure-based virtual screening uses docking engines and scoring functions to estimate binding modes and approximate affinities; modern deep learning approaches have introduced fast, learning-based pose prediction and generative docking which can both accelerate and sometimes improve pose quality relative to exhaustive search [22]. In addition, in silico ADMET and off-target prediction models are routinely applied to filter candidates for toxicity, pharmacokinetics, and polypharmacology risk. Combining these structure- and property-aware filters with transcriptome-derived target priorities enables a target-driven validation funnel for generated molecules [23]. 
There is growing interest in conditioning molecule design on biological context, but most prior efforts are limited to repurposing/selection of existing compounds based on signature matching, or population-level conditional generation. Direct generation of novel, patient-tailored chemotypes that are simultaneously conditioned on transcriptomic signatures, assembled by fragment-aware metaheuristics, and filtered against prioritized targets with structure-aware docking remains an open niche [23].
Taken together, the shows strong literature, independent progress in expression-based disease mapping and signature databases, generative chemistry, fragment and metaheuristic search strategies, and modern structure-aware filtering. However, these advances have largely been applied in disconnected workflows. The proposed approach explicitly bridges these fields by using patient transcriptomes to condition fragment assembly via new hybrid metaheuristic search while enforcing target-driven structural and ADMET filters an integration designed to produce AML-personalized candidate molecules with plausible binding modes and acceptable drug-like properties.
\\

\section{\textbf{Methodology}}

\subsection{\textbf{Data acquisition}}
To generate a personalised de novo drug design framework for AML in the method of SBDD, this work utilises bulk RNA sequencing data, a standard approach in precision oncology for capturing average gene expression across a mixed population of tumour cells. Bulk RNA-seq enables the identification of dysregulated genes, signalling pathways, and transcriptional signatures relevant to therapeutic targeting.
The proposed work is based on the Cancer Genome Atlas Program (TCGA) exon expression from the University of California, Santa Cruz (UCSC) genome characterisation centre at Xena Browser [24]. It allows scientists to explore functional genomic data sets for correlations between genomic and/or phenotypic variables. 
The primary dataset is from experimentally measured exon-level transcription estimates by RNAseq Illumina's HiSeq 2000 sequencing system, enabling individual labs to take on larger and more complex studies, including routine human genome sequencing with a High Accuracy and Unprecedented Output [25]. The dataset contains 173 samples, each represented as a column following the TCGA barcode format TCGA-SS-PPPP-TT. 

\begin{table}[!t]
\centering
{\footnotesize
\caption{\footnotesize Categorization and interpretation of gene expression levels based on normalized transcriptomic values} 
\label{turns_1}
\begin{tabular}{p{0.35\columnwidth} p{0.55\columnwidth}}
\toprule
Range/Value & Interpretation \\
\midrule
0 & No expression \\
1    & Very low expression \\
2–3 & Moderate expression \\
$>4$ & High expression \\
$>6$ & Very high expression \\
\bottomrule
\end{tabular}
}
\end{table}

Because the dataset is provided at the exon level, proper genomic mapping is essential to ensure that both the raw RPKM values and the processed $\log_2$ (RPKM+1) data can be accurately interpreted. Each row in the TCGA AML exon matrix corresponds not to a gene name but to a genomic coordinate. These coordinates must be mapped to their corresponding exon identifiers, gene symbols, and transcript annotations. To achieve this, the UCSC Xena platform uses a specialized probeMap  that links each genomic coordinate to its annotated exon within the hg19 reference genome. 
This mapping step ensures that raw level exon quantification files and their transformed data counterparts are aligned to the exact same exon definitions, preventing inconsistencies in annotation, misinterpretation of transcript structure, or errors in gene-level aggregation. Without this coordinate-to-exon mapping, the expression matrix would contain numerical values disconnected from their biological context, making it impossible to determine which genes or functional regions are dysregulated in AML. 

\subsection{\textbf{The Transcriptome-Conditioned De Novo Drug Generation Pipeline for AML}}

\begin{figure}[!t]
  \centering
  \includegraphics[width=1\linewidth]{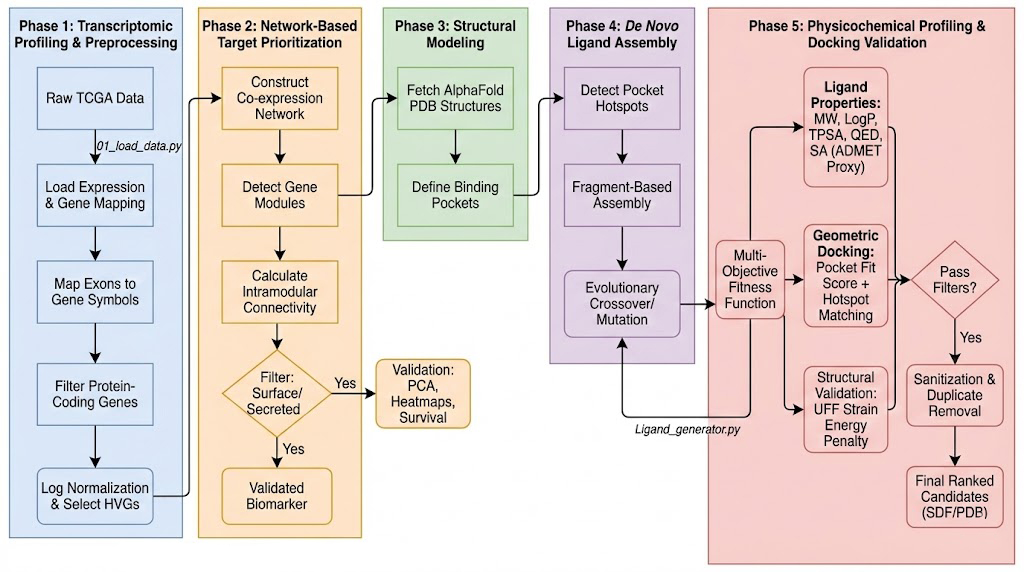}  
  \caption{An end-to-end in silico workflow for identifying surface-expressed biomarkers and optimizing candidate small-molecule inhibitors.}
  \label{fig:coffee_1}
\end{figure}

To translate raw transcriptomic data into personalized drug candidates tailored to specific molecular targets, this study developed a multi-stage computational pipeline. The workflow follows a sequential logic from data preprocessing to structural modeling and finally, evolutionary ligand design. The pipeline consists of five primary phases:
Phase (1): The objective of this phase is to reduce noise and isolate biologically relevant signals from the raw data. This phase includes exon-to-gene aggregation, protein-coding filtration, additional normalization, and feature selection. This transcriptomic preprocessing resulted in proteins designated as biomarkers (targets) for the future generation of the ligand.
Phase (2): Systems biology approaches are utilized in this phase to identify critical driver genes (biomarkers) rather than simple differential expression. The process includes co-expression network construction, module detection, hub gene identification, and targetability filtering. Network-based target discovery and prioritization are based on intramodular connectivity, involving various steps to result in valid biomarkers, specifically the top 20.
Phase (3): Once the biomarkers are prioritized, their physical structures the "locks" for the potential drugs must be modeled. This is a short but vital step in structural modeling, which includes the retrieval of target biomarkers using AlphaFold3 and the filtering out of unknown proteins via Quality Control (pLDDT).
Phase (4): Before ligand generation, the precise binding interface on the target protein is defined through critical and intensive work to detect binding pockets on the targets (hotspot mapping). This resulted in coordinates being extracted and saved in JSON metadata to coordinate systems to guide the ligand assembly process in the next phase.  
Phase (5): A custom-built Genetic Algorithm (GA) is utilized in this phase to evolve novel molecules that fit the Phase 4 pockets. A multi-objective fitness function is employed to explore the generated ligands and assign scores at different checkpoints based on Fragment-Based Assembly, resulting in top-generated candidate ligands in PDB and SDP formats. These ligands are subjected to different aspects of testing, including but not limited to ADMET properties, Molecular Weight (MW), and Topological Polar Surface Area (TPSA).
\subsection{\textbf{Transcriptomic Profiling \& Preprocessing}}
To generate robust molecular inputs for the downstream drug generation pipeline, the raw transcriptomic data underwent a rigorous three-step preprocessing protocol designed to aggregate genomic signals, filter for druggable targets, and reduce dimensionality.
\subsubsection{\textbf{Exon-to-Gene Aggregation}}

Raw exon-level expression data was first mapped to gene-level profiles to provide a biologically interpretable format. Using the UNC v2 exon-to-hg19 probe mapping reference, exon identifiers were aligned with their corresponding Human Genome Organisation (HUGO) gene symbols. To derive a single expression value for each gene, after computed the arithmetic mean of all exon probes associated with that specific gene locus. This step condensed the high-dimensional exon dataset into a gene-level expression matrix: $M_{\mathrm{genes}\times\mathrm{samples}}$
, ensuring that transcriptional activity was represented per functional gene unit.
\subsubsection{\textbf{Protein-Coding Biotype Filtration}}
It was essential to exclude non-coding RNA and pseudogenes. The metohed is utilized the "mygene" Python library to query the \texttt{type\_of\_gene} field for all gene symbols in the aggregated dataset against the human reference database. A strict inclusion filter was applied to retain only genes classified as "protein-coding." This filtration step ensures that all subsequent candidate biomarkers represent physically translatable structures capable of forming ligand-binding pockets.

\subsubsection{\textbf{Normalization and Highly Variable Gene (HVG) Selection}}
To account for the heteroscedasticity inherent in RNA-seq data and to focus the analysis on biologically distinct signals, the filtered dataset underwent normalization and feature selection:

\begin{figure}[!t]
  \centering
  \includegraphics[width=1\linewidth]{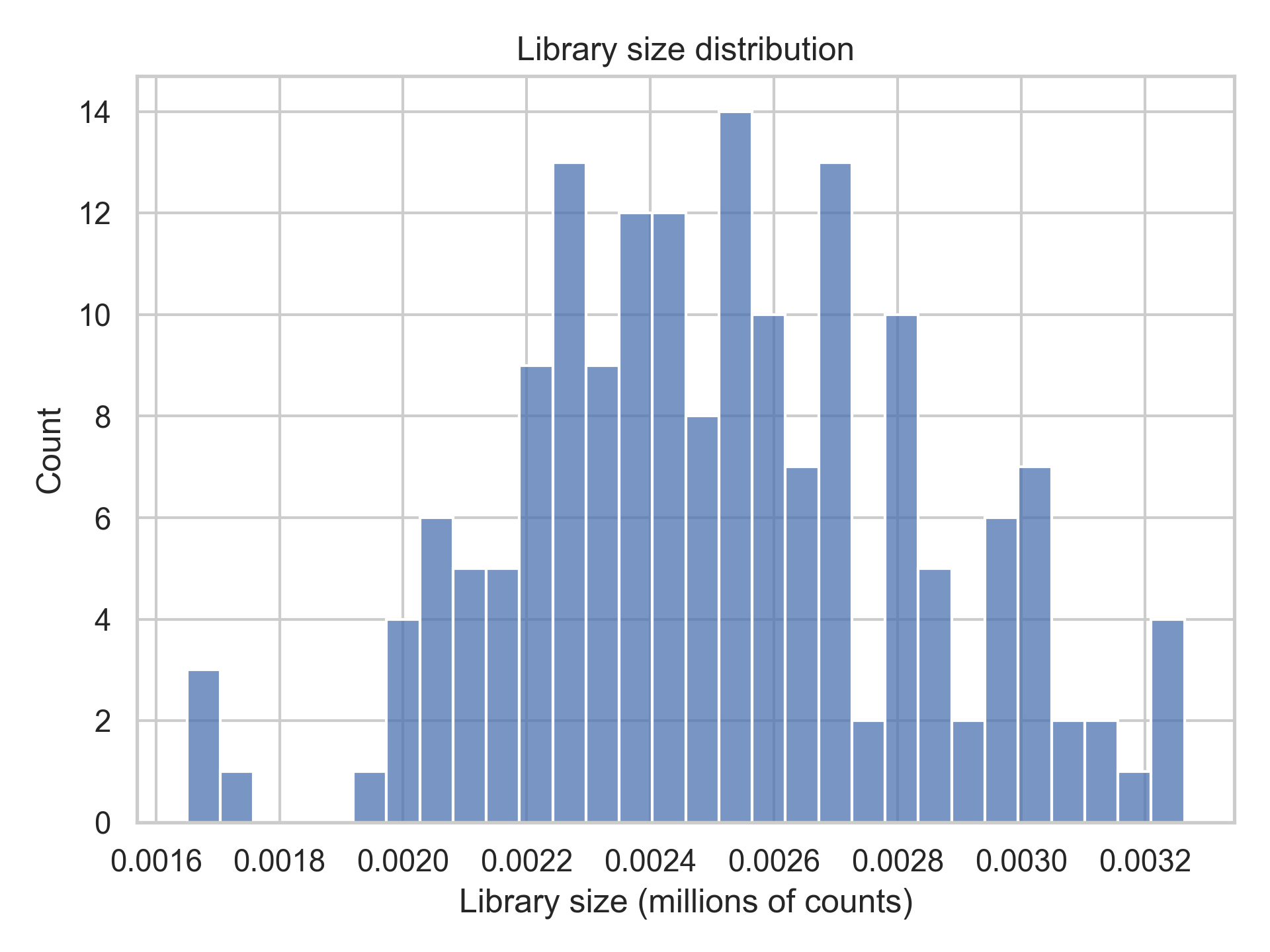}  
  \caption{Library sizes were calculated by summing read counts per sample; which the histogram displays a Gaussian-like distribution centered around the mean count depth. The absence of heavy tails indicates no failed libraries or significant outliers in sequence}
  \label{fig:coffee_2}
\end{figure}
The bimodal distribution is characteristic of high-quality data: the left peak represents low-abundance background noise, while the right peak represents actively expressed genes. The lack of deviant curves confirms the absence of major batch effects. Low-Expression Filtering is used to reduce technical noise, genes with a mean log-transformed expression value below a threshold of 0.5 across the cohort were discarded. Moreover, the HVG Selection is a primary step that prioritized genes contributing the most to patient heterogeneity. Variance was calculated for each gene across all samples. The genes were ranked by variance, and the top N=2,000 Highly Variable Genes (HVGs) were selected. These HVGs serve as the input for the Weighted Gene Co-expression Network Analysis (WGCNA) in Phase 2, ensuring the network is built on drivers of molecular variation rather than housekeeping genes. 

\begin{figure}[!t]
  \centering
  \includegraphics[width=1\linewidth]{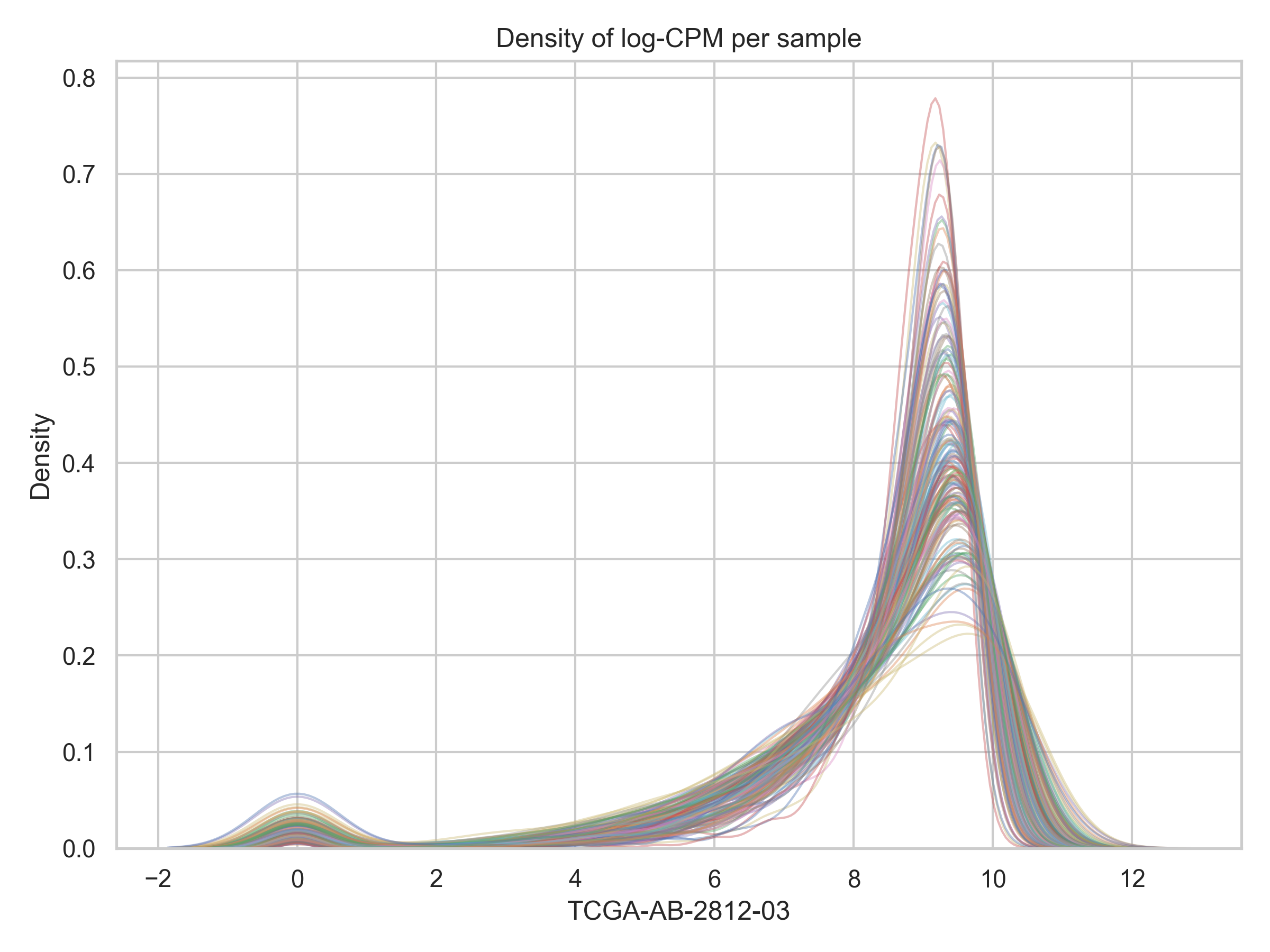}  
  \caption{The probability density function of gene expression was estimated using Kernel Density Estimation (KDE); which is the plot shows highly consistent expression profiles across all samples (overlapping curves)}
  \label{fig:coffee_3}
\end{figure}

\begin{figure}[!t]
  \centering
  \includegraphics[width=1\linewidth]{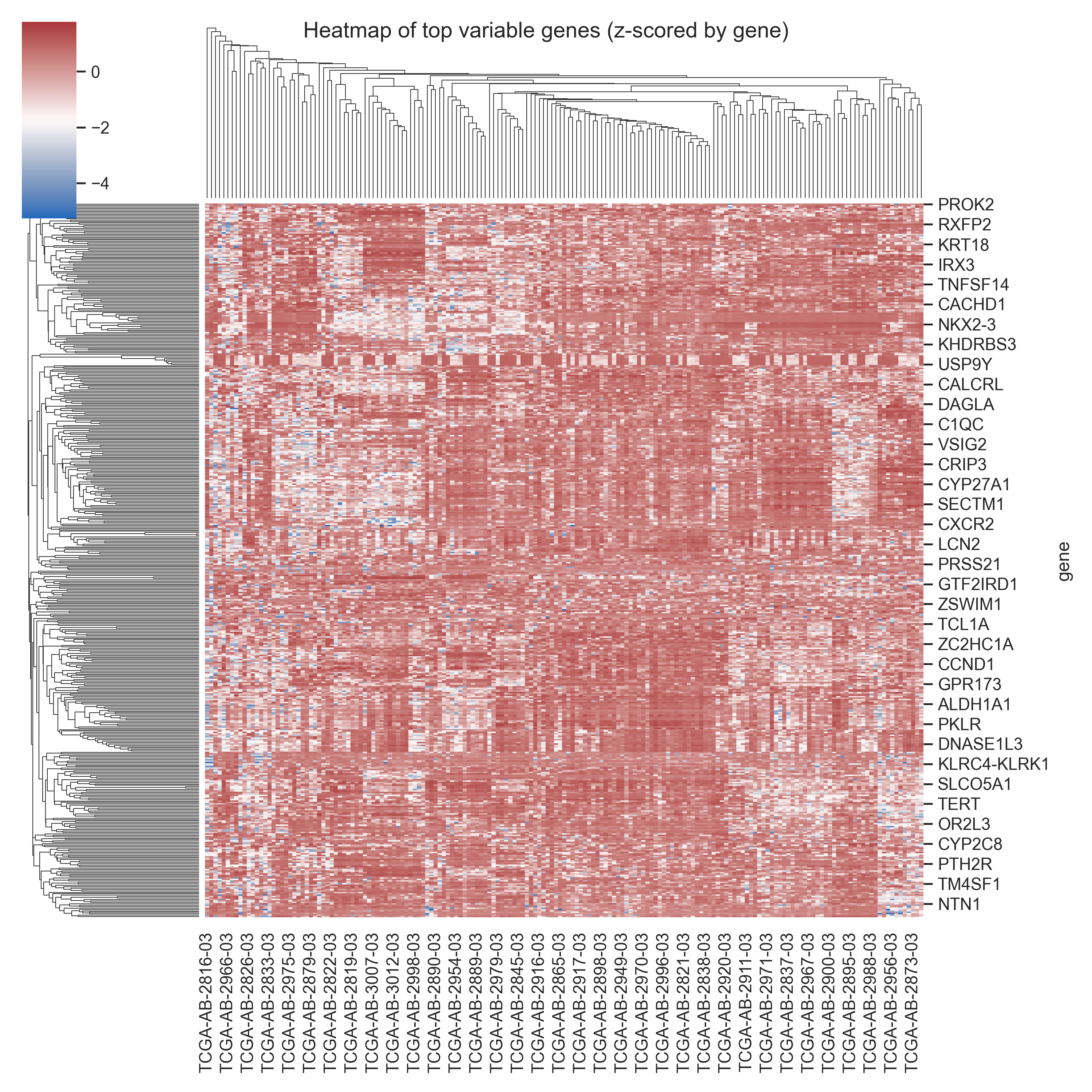}  
  \caption{Expression values of the top HVGs were Z-score normalized by gene and clustered using Euclidean distance and Ward’s linkage.}
  \label{fig:coffee_4}
\end{figure}
The heatmap in the figure 4 validates the selection process, revealing distinct gene modules. The "checkerboard" pattern indicates that these genes effectively capture the molecular heterogeneity necessary to distinguish potential subtypes within the dataset.
Following HVG selection, a validation process occurred in the data structure using linear and non-linear dimensionality reduction techniques to ensure the dataset contained discernable biological signals as.

\begin{figure}[!t]
  \centering
  \includegraphics[width=1\linewidth]{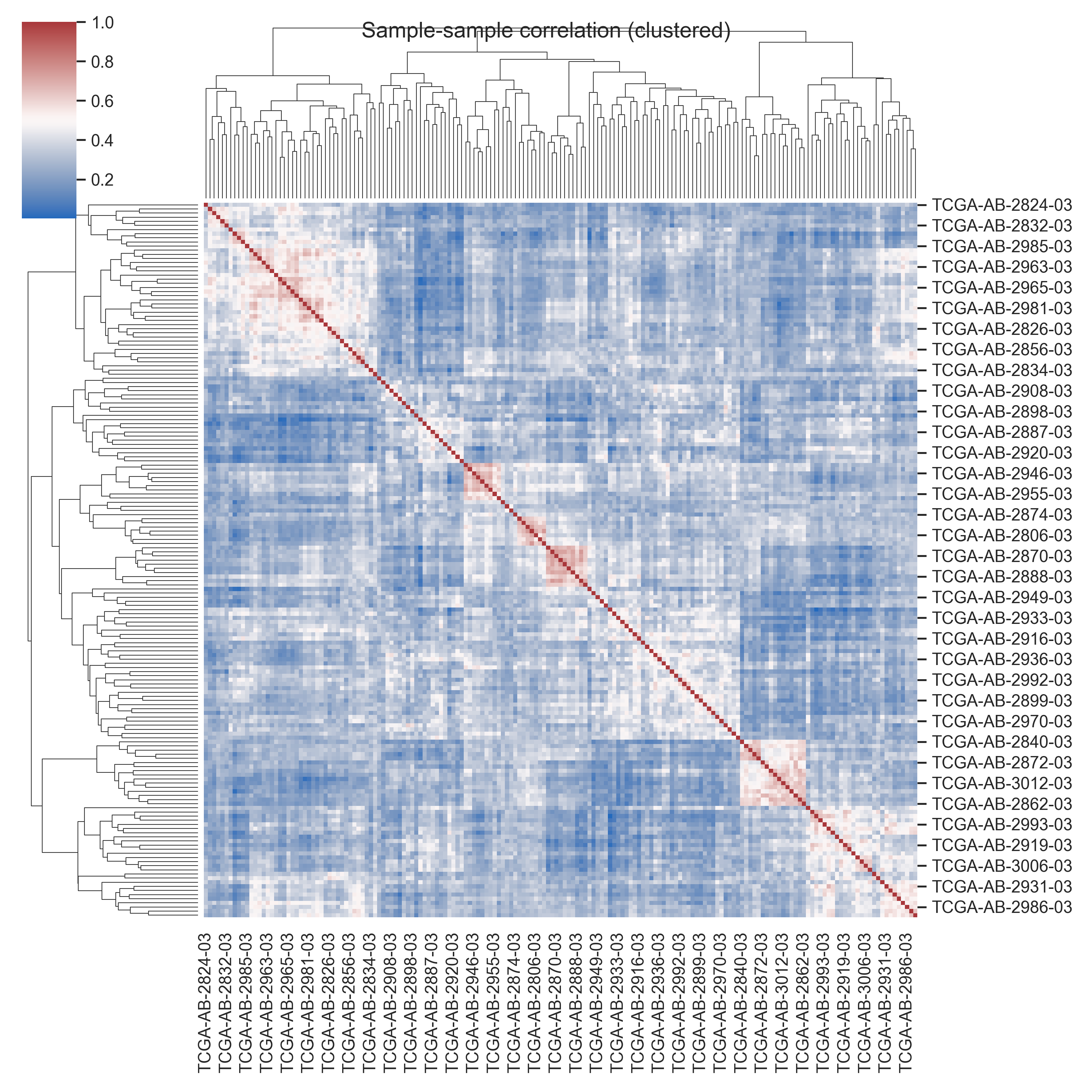}  
  \caption{Sample-Sample Correlation Matrix}
  \label{fig:coffee_5}
\end{figure}
The pairwise Pearson correlation matrix reveals low global correlation blue areas, confirming biological heterogeneity. However, distinct blocks of high correlation by red diagonal identify cliques of samples with similar transcriptional profiles, suggesting the presence of molecular subtypes.

\begin{figure}[!t]
  \centering
  \includegraphics[width=1\linewidth]{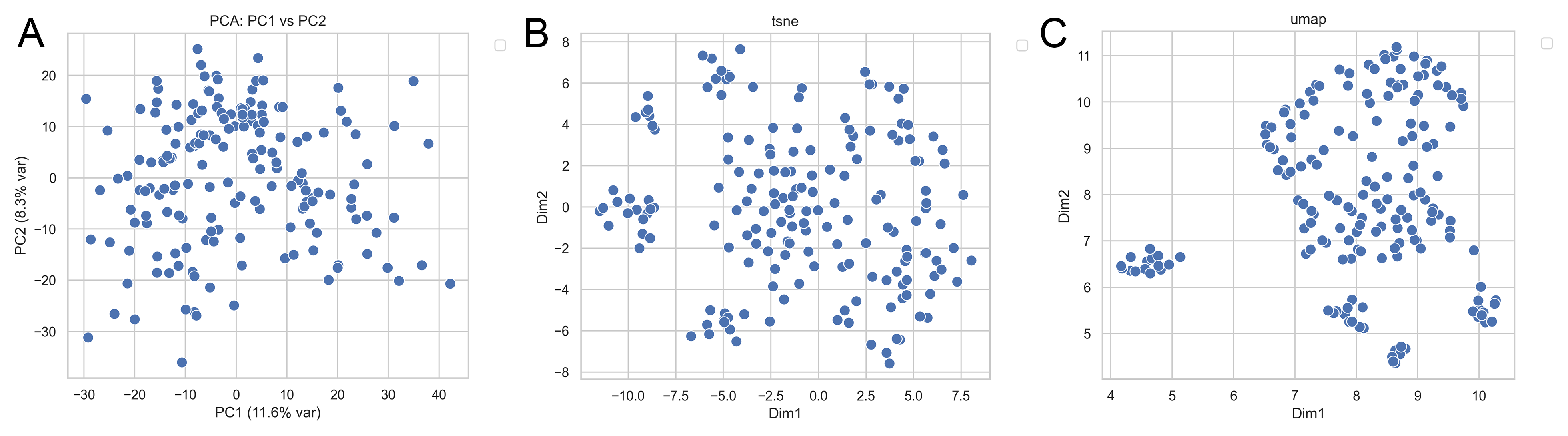}  
  \caption{Dimensionality Reduction PCA, t-SNE, and UMAP: Data was projected into 2D space using PCA (linear), t-SNE, and UMAP (non-linear).}
  \label{fig:coffee_6}
\end{figure}

PCA at Figure~6A reveals that the top two components explain $\sim$20\% of the variance, indicating a complex, high-dimensional dataset. Both t-SNE and UMAP at Figure~6A~\&~6B show a largely continuous distribution of samples rather than discrete, separated clusters, though a small, distinct island of samples is visible in the non-linear projections, potentially representing a rare subtype.

\begin{figure}[!t]
  \centering
  \includegraphics[width=1\linewidth]{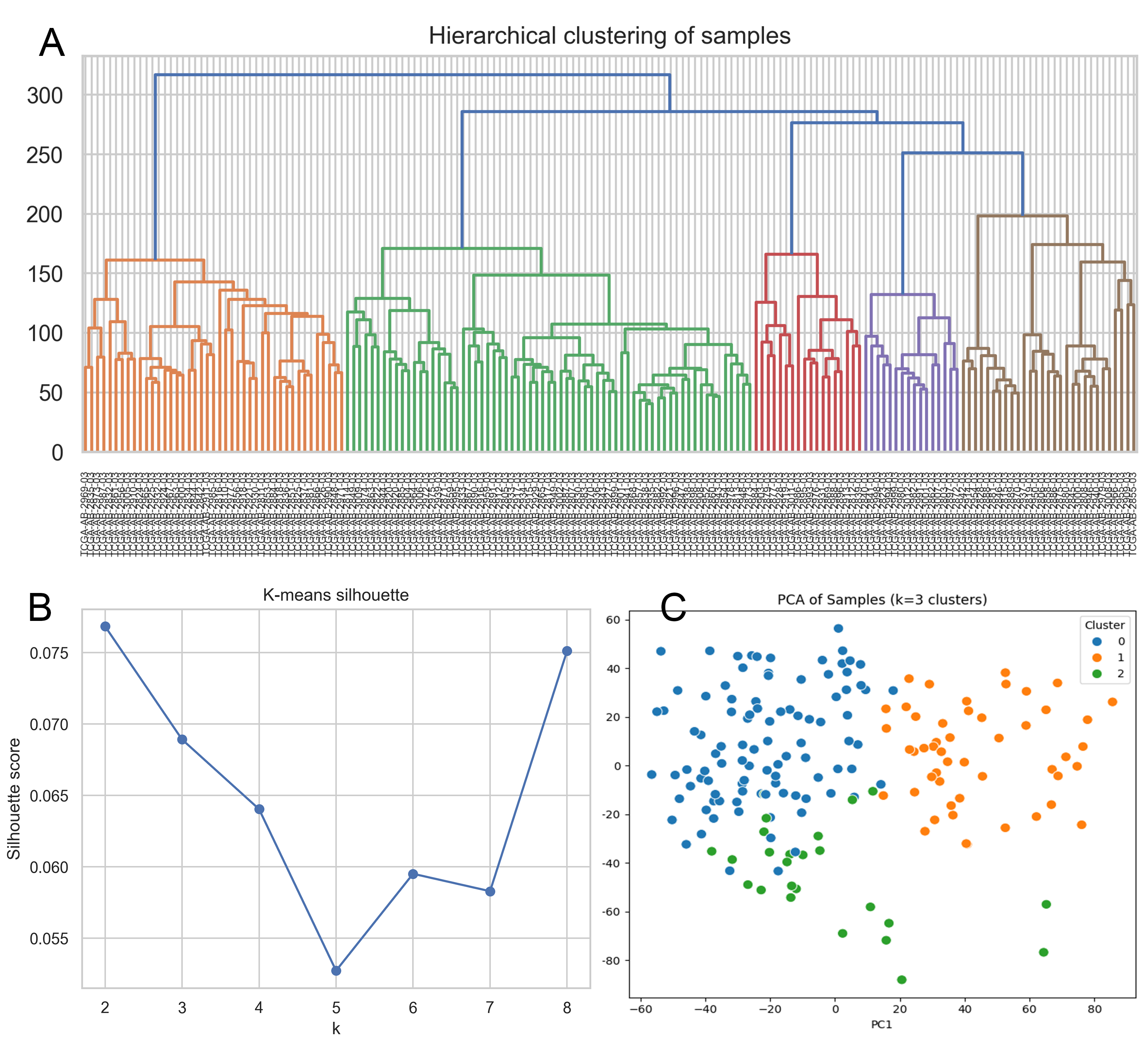}  
  \caption{Clustering Analysis where K-means clustering was assessed via Silhouette scores, and hierarchical clustering was visualized via a dendrogram.}
  \label{fig:coffee_7}
\end{figure}

The Silhouette analysis at figure 7B shows generally low scores ($<$0.08), consistent with the continuous nature of the data seen in PCA. While K-means K=3 partitions the PCA space as shown in figure 7C the boundaries are not strict. 
The Hierarchical Dendrogram as shown in figure 7A identifies roughly five major clades, suggesting that while the data is continuous, it possesses a hierarchical taxonomy of subtypes that can be exploited for the AML drug target discovery.

\subsection{\textbf{Weighted Gene Co-expression Network Analysis (WGCNA)}}
\subsubsection{\textbf{Rationale and Network Construction}}

The objective of this analysis was to transform raw gene expression profiles into a systems-level understanding of gene–gene interactions in AML, thereby enabling the identification of AML-specific co-expression modules, biomarker candidates, and putative therapeutic targets. Co-expression network analysis was selected because it captures higher-order transcriptional organization and coordinated gene activity that cannot be detected through single-gene differential analysis alone.

A weighted gene co-expression network was constructed using the 2000 HVGs selected in Phase~1. First, a symmetric gene--gene Pearson correlation matrix was computed across 173 AML samples, capturing linear co-expression patterns:
\begin{itemize}
  \item High positive correlation indicates genes that exhibit coordinated behavior across AML patients, often reflecting shared pathway involvement.
  \item Low or near-zero correlation suggests independent or uncoordinated transcriptional activity.
\end{itemize}

To enhance meaningful biological signals while suppressing noise, the correlation matrix was transformed into a weighted adjacency matrix using a soft-thresholding power function, as shown in Equation~(1).

\begin{equation}
A_{ij} = \lvert r_{ij} \rvert^{\beta}
\end{equation}

\begin{figure}[!t]
  \centering
  \includegraphics[width=1\linewidth]{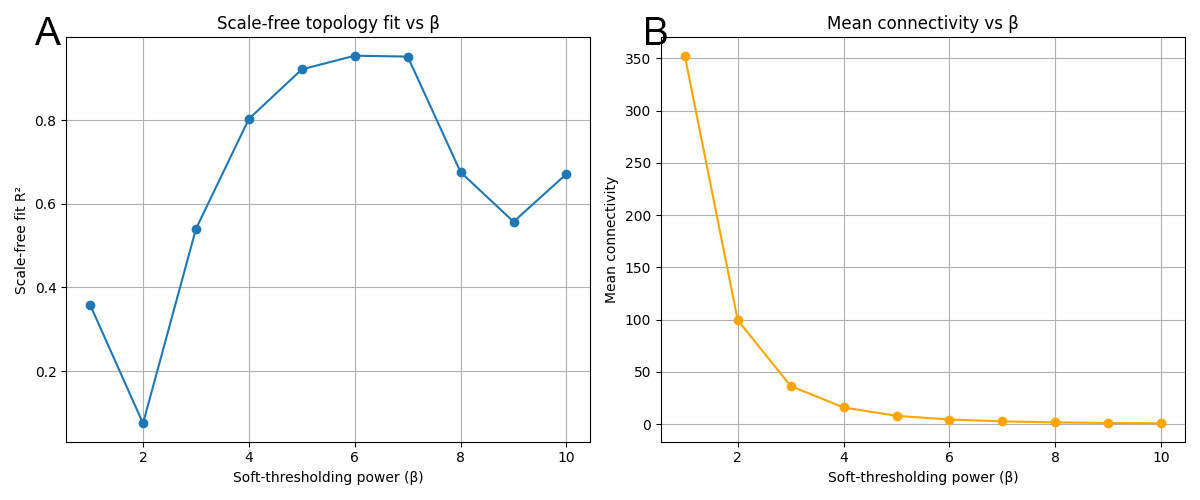}  
  \caption{Analysis of Network Topology for Various Soft-Thresholding Powers}
  \label{fig:coffee_8}
\end{figure}

With $\beta = 6$ selected to approximate a scale-free topology based on Fig.~8A~\&~8B. 
This value represents the inflection point that maximizes the scale-free topology fit 
($R^2 > 0.85$) while maintaining sufficient mean connectivity to ensure the network 
remains biologically interconnected and not overly sparse. 
This step mirrors the core principle of WGCNA, ensuring that strong gene--gene 
relationships contribute disproportionately to module formation. 
Self-loops were removed by setting all diagonal entries to zero.

\subsubsection{\textbf{Module Characterization and Hub Gene Identification}}

Each module was summarized by its corresponding module of eigengene, defined as the first principal component (PC1) of the module’s normalized expression matrix. For modules containing at least two genes, eigengenes were computed using PCA and served as representative expression signatures that capture dominant transcriptional variation within each module.

\begin{figure}[!t]
  \centering
  \includegraphics[width=1\linewidth]{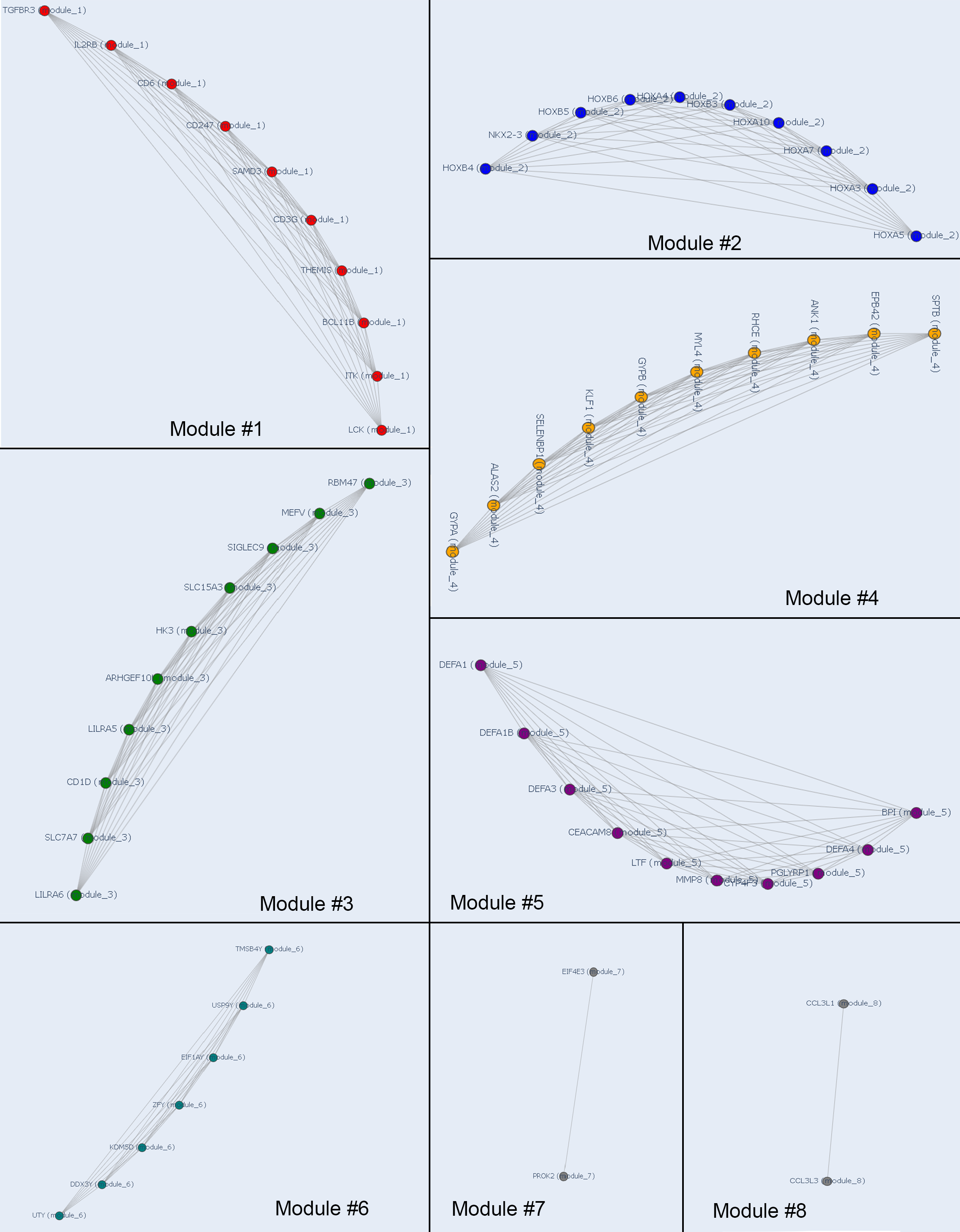}  
  \caption{Global Topology of the AML Transcriptome intramodular via Chord Diagram}
  \label{fig:coffee_9}
\end{figure}

To interrogate the systemic relationships between the eight identified co-expression modules, the constructed Module Eigengene Network was implemented. For each module, a "Module Eigengene" (ME) defined as the first principal component of the module's expression matrix was calculated to represent its summary expression profile. Then computed the Pearson correlation matrix between all pairwise MEs.
Intramodular hub genes were then identified by quantifying intramodular connectivity, where genes with the highest summed edge weights within a module were classified as hubs. Hub genes are particularly important because, with a total of 8 modules that shown at figure 9,

\begin{table}[!t]
\centering
\footnotesize
\caption{Criteria for identifying candidate therapeutic targets through intramodular connectivity and hub gene analysis}
\label{turns_2}
\begin{tabular}{p{0.32\linewidth} p{0.6\linewidth}}
\toprule
Co-expression Result & Interpretation \\
\midrule

HVGs $\rightarrow$ modules 
& Select biologically meaningful gene sets; reduce noise and focus on AML-relevant patterns. \\

Module eigengene (PC1) 
& Summarizes module expression and is used to track overall module activity across samples. \\

Intramodular connectivity ($k_{\text{Within}}$) 
& Identifies strong biomarker candidates that act as key regulators of AML transcriptional networks. \\

Hub genes 
& Quantifies gene centrality; highly connected genes are prioritized as therapeutic targets. \\
\bottomrule
\end{tabular}
\end{table}

\subsection{\textbf{Identification and Characterization of Top 20 Biomarkers in AML}}

\subsubsection{\textbf{Rationale and Selection of Biomarkers}}

While the preceding co-expression network analysis considered the broader landscape of highly variable genes, this step focused on narrowing the gene set to a manageable subset of highly informative candidates. To achieve this, a ranking of WGCNA selection was employed, leveraging the assumption that genes exhibiting the largest variance across AML patients are most likely to reflect meaningful biological differences rather than stochastic noise. Hemoglobin-related genes were excluded from consideration, as their high expression is often non-specific and can obscure the detection of true AML-relevant signals. Following this filtering process, the top 20 genes were retained as candidate biomarkers, representing a balance between variability and biological relevance.

\begin{table}[!t]
\centering
\footnotesize
\caption{Top 20 candidate biomarkers selected from the WGCNA-ranked gene list after excluding hemoglobin-related genes}
\label{turns_3}
\begin{tabular}{p{0.15\linewidth} p{0.30\linewidth} p{0.45\linewidth}}
\toprule
Biomarker & Targetability & Function \& AML Relevance \\
\midrule
S100A9 & Secreted Alarmins \& Inflammatory Mediators
& Pro-inflammatory alarmin highly expressed in AML (M4/M5 subtypes). \\

HK3 & Intracellular Enzymes \& Metabolic Regulators
& Myeloid-specific hexokinase; overexpression drives increased glycolytic flux. \\

SLC7A7 & Surface / Secreted
& Cationic amino acid transporter (y+L system); critical for arginine uptake. \\

RBM47 & Intracellular Metabolic / Signaling
& RNA-binding protein involved in mRNA stability; regulates myeloid differentiation. \\

MEFV & Intracellular Enzymes \& Metabolic Regulators
& Encodes Pyrin, a regulator of the inflammasome; enriched in monocytic AML. \\

LILRA5 & Surface Receptors \& Transporters
& Innate immune receptor; activation triggers pro-inflammatory cytokine release. \\

SIGLEC9 & Surface Receptors \& Transporters
& Immunosuppressive receptor inhibiting NK and T-cell cytotoxicity. \\

CD300E & Surface Receptors \& Transporters
& Immune-activating receptor expressed on myeloid cells. \\

LILRA6 & Surface Receptors \& Transporters
& Immunoglobulin-like receptor regulating immune activation. \\

FGD2 & Intracellular Enzymes \& Metabolic Regulators
& CDC42-specific GEF; regulates actin cytoskeleton remodeling. \\

VCAN & Secreted Alarmins \& Inflammatory Mediators
& ECM proteoglycan promoting blast motility and leukemic niche formation. \\

FGL2 & Secreted Alarmins \& Inflammatory Mediators
& Immunomodulatory protein promoting T-cell suppression and coagulation. \\

SLC11A1 & Surface Receptors \& Transporters
& Divalent metal ion transporter regulating iron homeostasis. \\

C5AR1 & Surface Receptors \& Transporters
& Complement receptor promoting immunosuppressive microenvironment. \\

S100A12 & Secreted Alarmins \& Inflammatory Mediators
& RAGE ligand mediating chronic inflammation in AML. \\

NCF2 & Intracellular Enzymes \& Metabolic Regulators
& NADPH oxidase subunit regulating ROS production. \\

NCF1 & Intracellular Enzymes \& Metabolic Regulators
& NADPH oxidase cytosolic subunit enabling superoxide generation. \\

CD1D & Surface Receptors \& Transporters
& Lipid antigen-presenting molecule regulating iNKT cell responses. \\
\bottomrule
\hline
\end{tabular}
\end{table}

The top 20 prioritized biomarkers reveal a distinct targetable landscape characterized by the upregulation of myeloid-lineage surface receptors, metabolic transporters, and secreted inflammatory mediators. The analysis identified three primary functional clusters suitable for therapeutic intervention.
First, Solute Carrier (SLC) Transporters such as SLC7A7, SLC11A1, SLC15A3 and the metabolic enzyme HK3 represent high-value targets for small-molecule inhibition. The upregulation of the cationic amino acid transporter SLC7A7 suggests a metabolic dependency on exogenous arginine uptake to fuel leukemic blast proliferation, a vulnerability that can be exploited by starvation strategies or high-affinity competitive inhibitors. Similarly, HK3 overexpression underscores the 'Warburg effect' in AML, offering a defined enzymatic pocket for direct pharmacological blockade.
Second, the immune-modulatory surface receptors SIGLEC9, LILRA5/6, and CD300E highlight mechanisms of immune evasion. SIGLEC9, an inhibitory checkpoint receptor, recruits phosphatases to dampen NK and T-cell cytotoxicity; blocking its ligand-binding interface could restore anti-leukemic immunity. These surface-accessible proteins are optimal candidates for both small-molecule modulation (targeting the sialic acid binding cleft) and antibody-drug conjugate (ADC) delivery.
Finally, the secretome is dominated by the S100 Alarmin Family such as S100A8, S100A9, S100A12, and FGL2. These factors condition the bone marrow niche to support chemoresistance and suppress T-cell function. While traditionally targeted by neutralizing antibodies, the structural analysis suggests that their oligomerization interfaces and receptor-binding domains possess druggable hotspots suitable for novel protein-protein interaction (PPI) inhibitors generated by the proposed  de novo pipeline.

\subsubsection{\textbf{Exploratory Analysis of Expression Patterns}}

Hierarchical clustering was then applied to the top 20 genes to assess co-expression patterns and potential subgroupings among patients. A clustered heatmap was generated, which visualized both gene-gene relationships and patient clustering simultaneously. This approach allowed the identification of co-expressed gene clusters, reflecting putative functional modules within this reduced biomarker set. 

\begin{figure}[!t]
  \centering
  \includegraphics[width=1\linewidth]{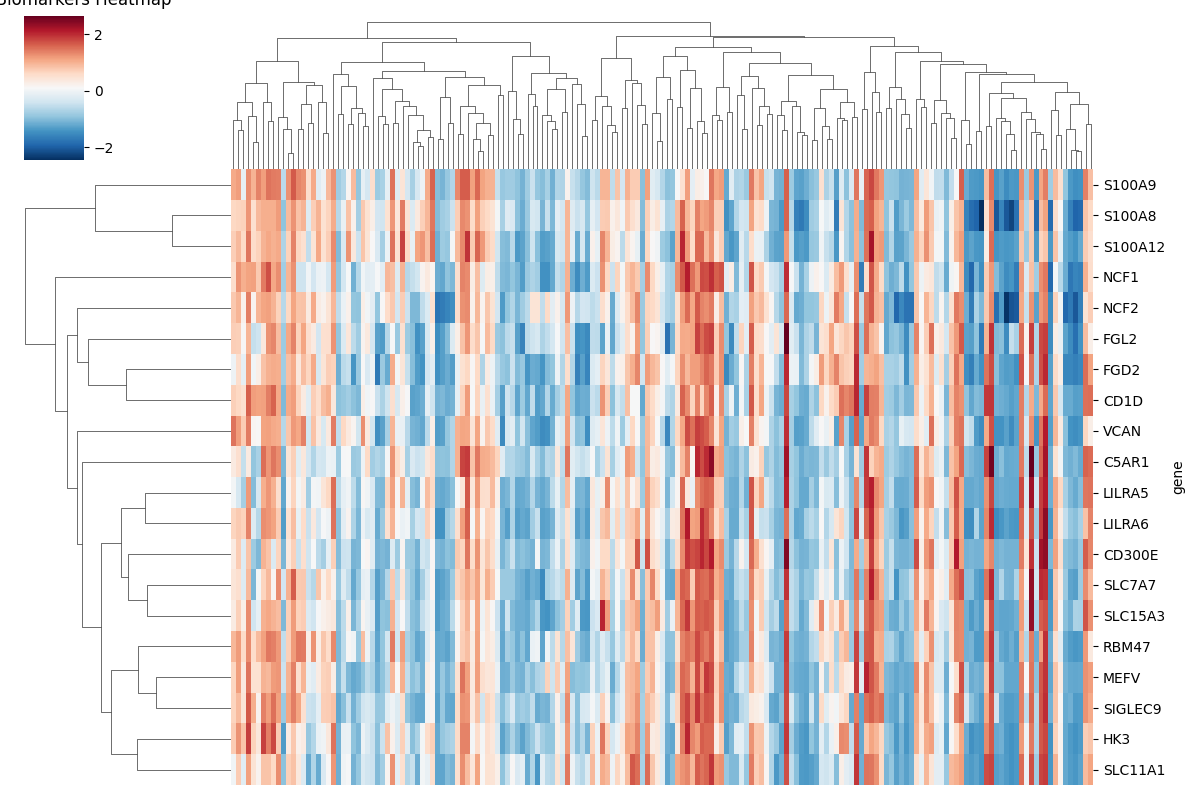}  
  \caption{Hierarchical clustering of the transcriptomic landscape of prioritized AML biomarkers}
  \label{fig:coffee_10}
\end{figure}

To elucidate the systemic interrelationships among the top 20 biomarkers, using a constructed undirected weighted co-expression network based on pairwise Pearson correlations. Only gene pairs exceeding a correlation threshold of 0.7 were connected, with edge weights reflecting co-expression strength. This network topology revealed hub genes with highly connected nodes likely to act as central drivers of the identified AML transcriptional program. Additionally, the network structure uncovered distinct modular subgroups, suggesting shared regulatory mechanisms or functional pathways suitable for coordinated therapeutic targeting.
The heatmap, combined with hierarchical clustering, reveals sample-level patternsand identifies potential biomarker subgroups. Dendrograms, computed using Euclidean distance and average linkage, offer structural insight into how expression profiles cluster together, indicating whether biomarkers define coherent biological modules.

\begin{figure}[!t]
  \centering
  \includegraphics[width=1\linewidth]{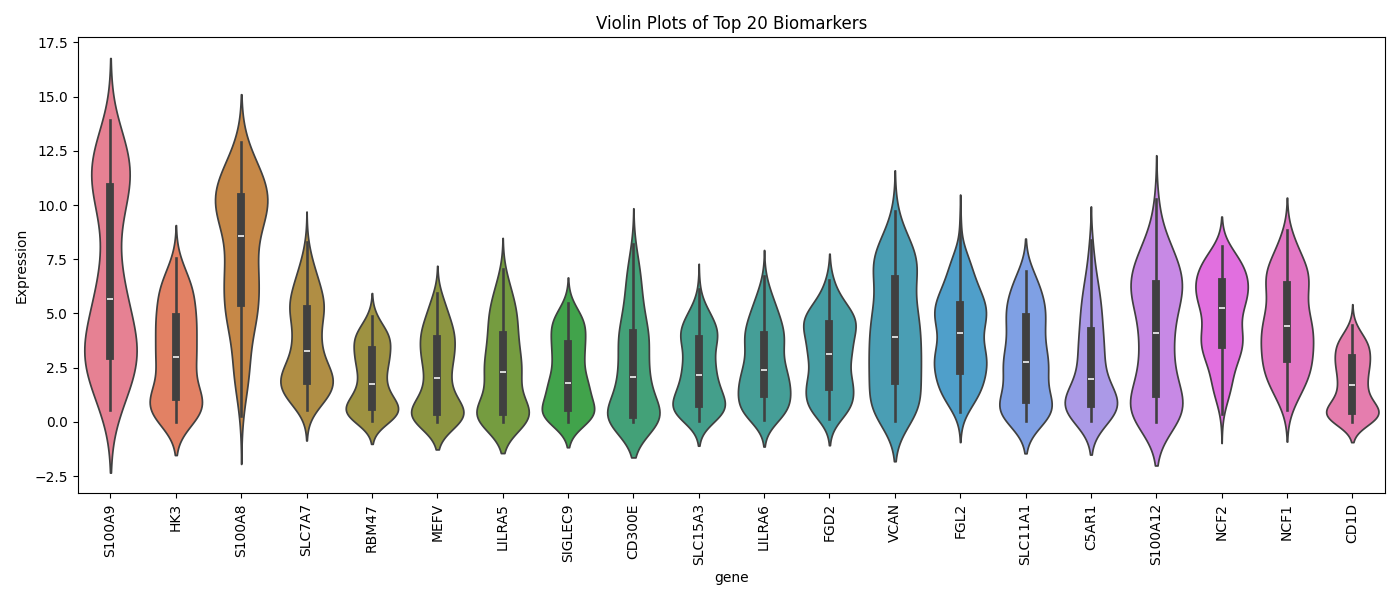}  
  \caption{HViolin plots illustrating the full expression density profiles of the top 20 biomarkers. The plot emphasizes multimodal and asymmetric expression patterns in several genes, particularly the highly expressed S100-family markers, and reveals clear separation between high-, medium-, and low-expression genes. These density shapes provide additional evidence of the biomarkers’ discriminatory potential across AML samples}
  \label{fig:coffee_11}
\end{figure}

\begin{figure}[!t]
  \centering
  \includegraphics[width=1\linewidth]{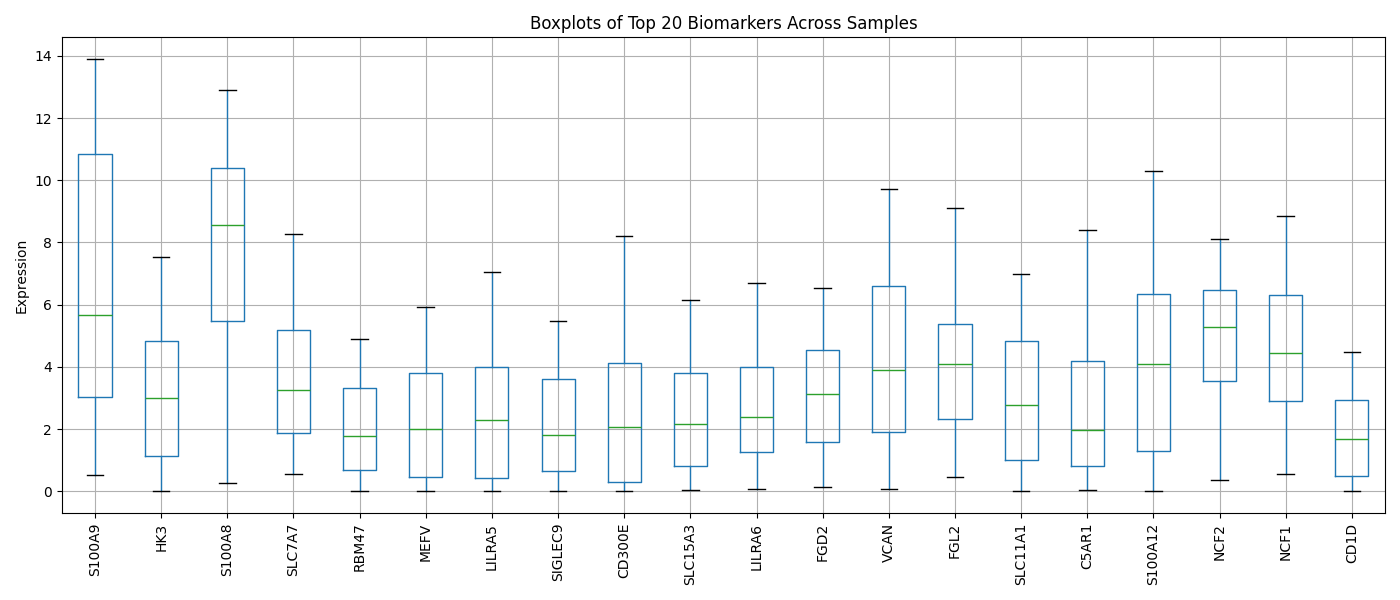}  
  \caption{Boxplots showing the expression distributions of the top 20 selected biomarkers across AML samples. The plots highlight substantial variability among genes, with several biomarkers exhibiting markedly higher median expression and wider dispersion, while others display tighter, lower-expression ranges. These distributional differences support their potential to capture heterogeneous AML-related expression patterns.}
  \label{fig:coffee_12}
\end{figure}

\begin{figure}[!t]
  \centering
  \includegraphics[width=1\linewidth]{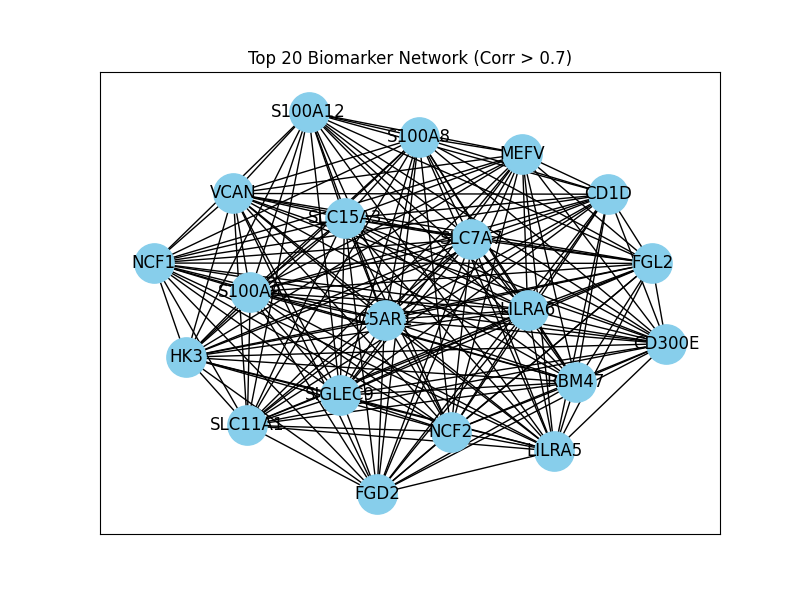}  
  \caption{Weighted Gene Co-expression Network of Prioritized Biomarkers}
  \label{fig:coffee_13}
\end{figure}

\begin{figure}[!t]
  \centering
  \includegraphics[width=1\linewidth]{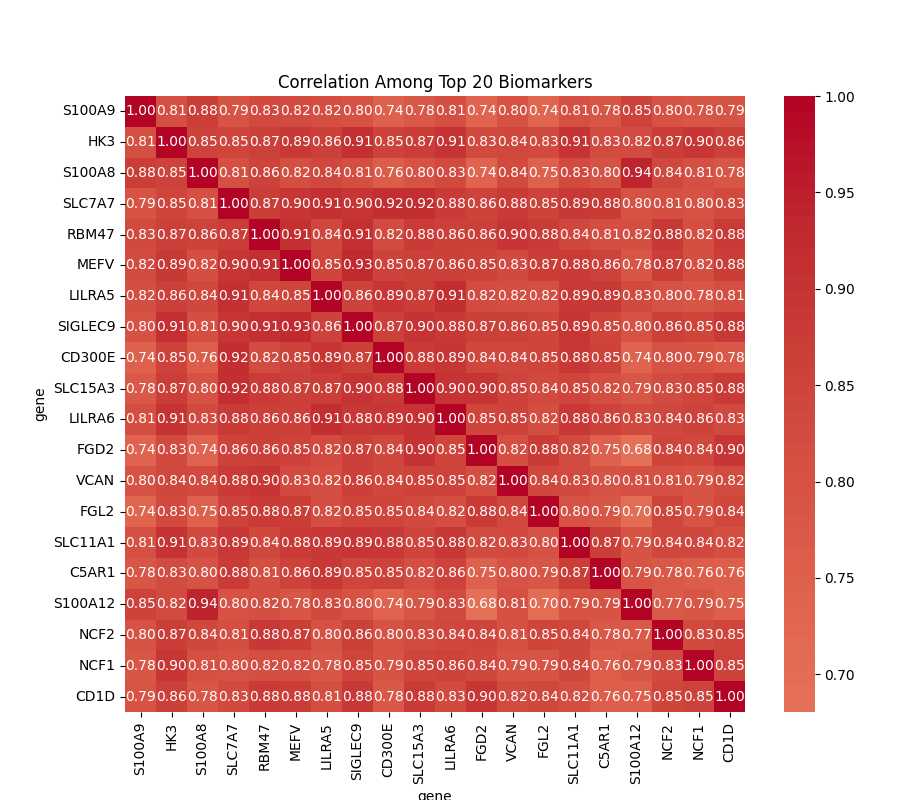}  
  \caption{Unsupervised hierarchical clustering heatmap of therapeutic targets}
  \label{fig:coffee_14}
\end{figure}

\begin{figure}[!t]
  \centering
  \includegraphics[width=1\linewidth]{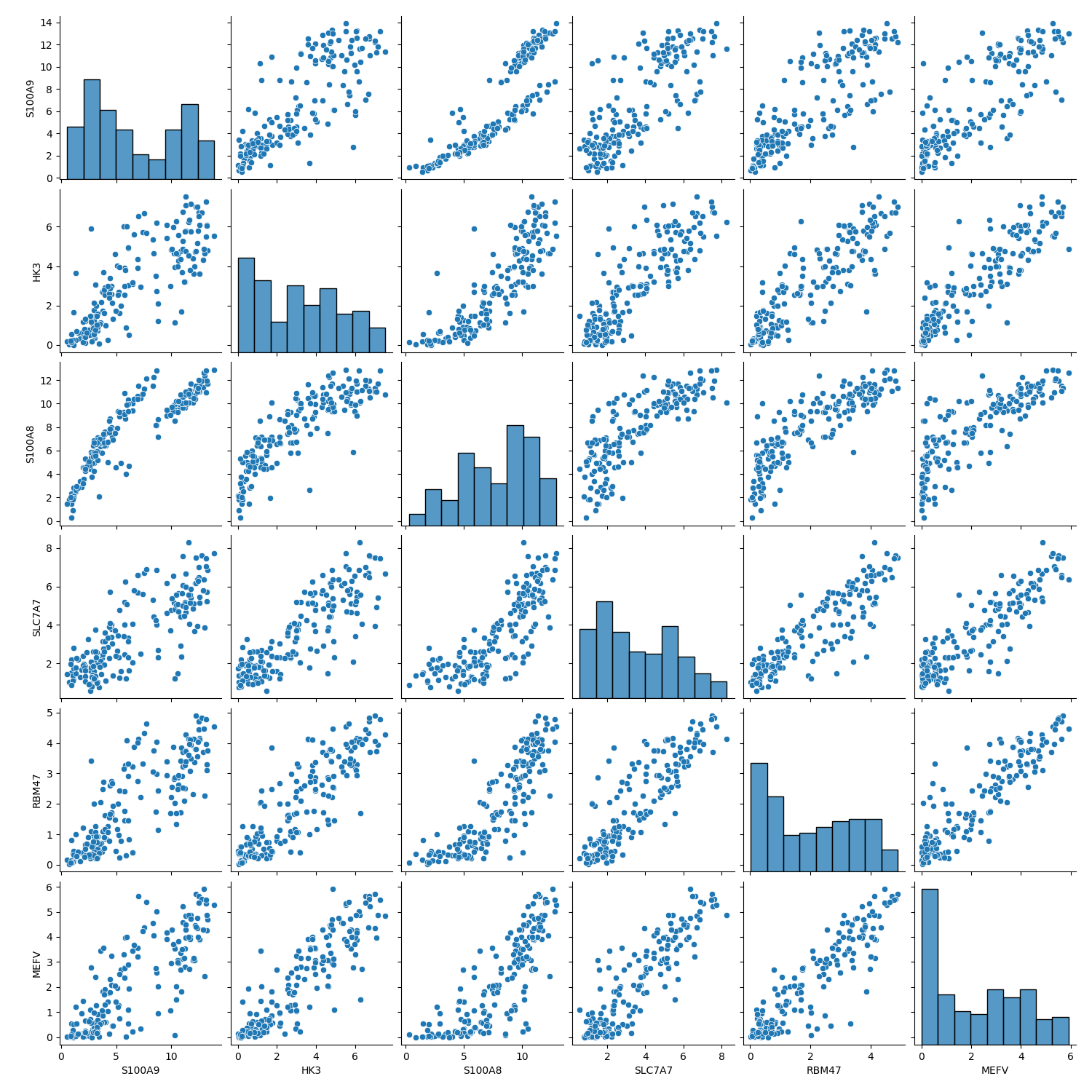}  
  \caption{Dimensionality Reduction Representation of the AML Cohort using some of the Top 20 Biomarkers}
  \label{fig:coffee_15}
\end{figure}

\subsubsection{\textbf{Structural Retrieval and Integration Using AlphaFold Models}}    
UniProt accessions were used to fetch predicted 3D structures from the AlphaFold3 Protein Structure Database [26]. HTTP requests were used to download each structure, and all successfully retrieved PDB files were stored in a structured directory for later inspection.
VCAN biomarker logged and flagged for manual examination because it doesn't have an available AlphaFold pdb strucutre. For those with existing renderings in the project directory, all images were assembled into a unified figure, ensuring that structural data could be directly visualized downstream.

\begin{figure}[!t]
  \centering
  \includegraphics[width=1\linewidth]{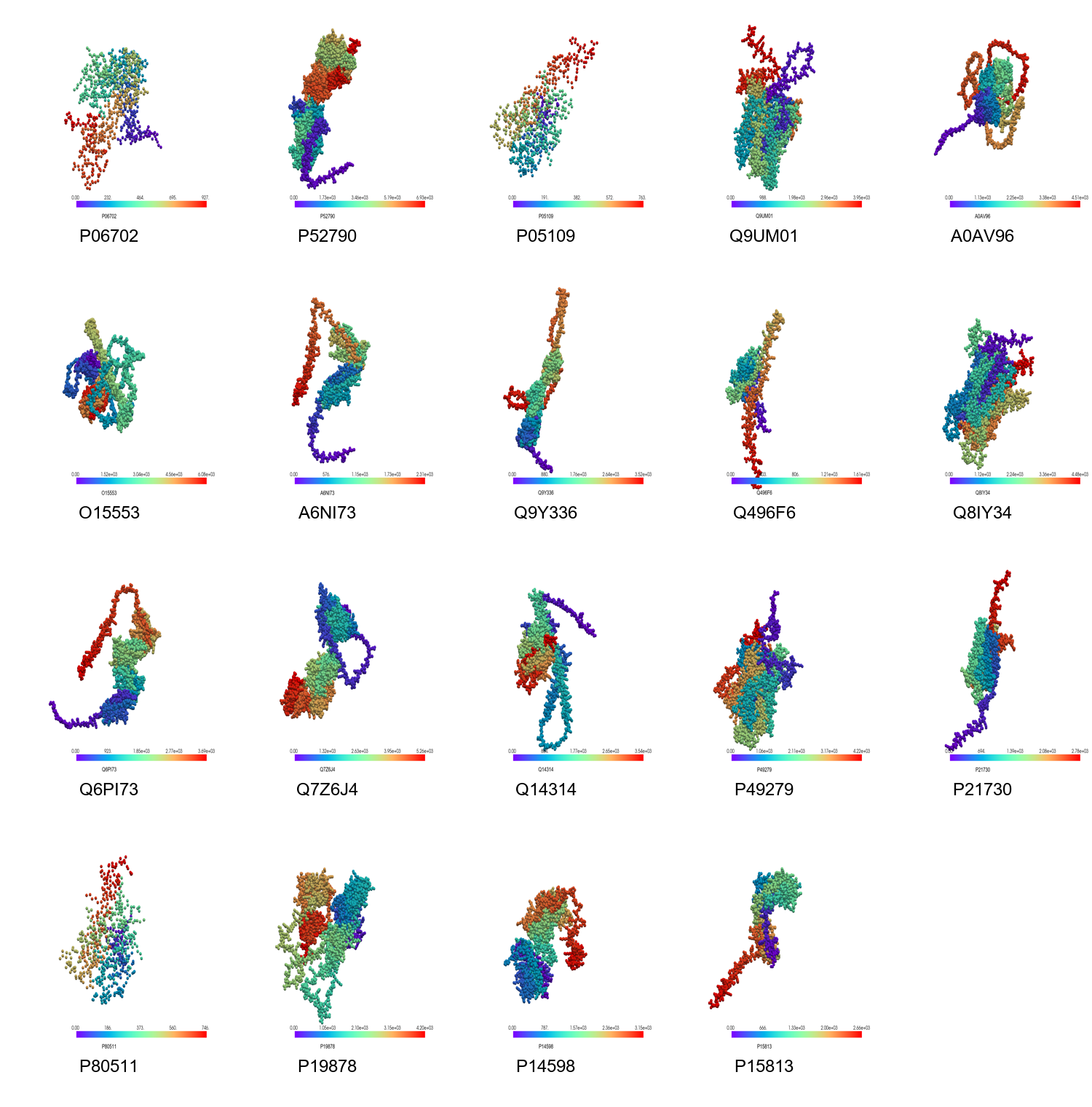}  
  \caption{Visualization of retrieved AlphaFold3 final top 19 biomarkers after excluding VCAN biomarker}
  \label{fig:coffee_16}
\end{figure}

Finally, all computed metrics values, composite scores, functional annotations, PCA coordinates, clustering results, correlation matrices, and structural availability information were merged into a single organized Excel workbook. This consolidation step ensures that researchers can navigate seamlessly from statistical ranking to biological interpretation and structural modeling, forming a complete biomarker discovery framework applicable to AML.

\subsubsection{\textbf{Geometric and Physicochemical Pocket Profiling with DOGSite}}    
To identify and characterize druggable pockets within the target protein, we used the DOGSiteScorer engine provided through the ProteinsPlus API [27]. The workflow consists of four major stages: structure upload, job execution, pocket detection, and feature extraction/scoring.
The protein target structure in PDB format was submitted to the DOGSite endpoint of the ProteinsPlus API. Upon submission, the service returned a unique \texttt{job\_id} corresponding to a server-side processing task. This asynchronous workflow ensures that binding pocket detection is completed prior to downstream analysis.

DOGSite internally performs grid-based geometric and physicochemical analysis on the protein surface to identify concave regions that exhibit cavity-like properties. The algorithm examines:

\begin{itemize}
  \item 3D surface topology.
  \item Local curvature.
  \item Enclosed volume regions.
  \item Spatial arrangement of residues and atoms.
\end{itemize}
After detection, pocket-level physicochemical and geometric features are retrieved with each entry including the following quantitative descriptors, which are central for ligandability assessment.
\FloatBarrier

\begin{table}[!t]
\centering
\footnotesize
\caption{Structural and physicochemical parameters used for the characterization and evaluation of binding pocket druggability.}
\label{turns_4}
\begin{tabular}{p{0.30\columnwidth} p{0.20\columnwidth} p{0.35\columnwidth}}
\toprule
Geometric Properties & Physicochemical Properties & Residue/Atom-Level Information \\
\midrule
Volume (Å³): size of the pocket cavity & Hydrophobicity: proportion of hydrophobic surface area within the pocket.
& Residues lining the pocket \\

Depth: average depth measured from the surface to the pocket core. & Aromaticity: prevalence of aromatic residues contributing to \texttt{$\pi$-$\pi$} interactions.
& 3D coordinates of pocket atoms \\

Enclosure (\%): degree of pocket “closedness,” reflecting how well the cavity is shielded from solvent. & H-bond donors and acceptors: counts of donor and acceptor atoms, indicating potential hydrogen-bonding capability.
& Geometry files for visualization \\

\bottomrule
\hline
\end{tabular}
\end{table}

These features collectively represent the binding potential of the pocket as shown in table IV.
To prioritize the most promising pockets, a composite scoring function was applied:
\begin{IEEEeqnarray}{rCl}
\text{Score} &=& 0.3 \times \text{Volume} 
               + 0.2 \times \text{Depth} 
               + 0.2 \times \text{Enclosure} \times 100 \nonumber\\
             && + 0.1 \times \text{Hydrophobicity} \times 100 \nonumber\\
             && + 0.1 \times \text{Aromaticity} 
               + 0.1 \times (\text{Donors} + \text{Acceptors})
\end{IEEEeqnarray}

This scoring formula balances geometric enclosure (important for ligand binding), cavity size, and chemical interaction potential.
After scoring:
\begin{itemize}

  \item Scores were normalized to a 0-1 scale.
  \item Pockets were sorted in descending order.
  \item Enclosed volume regions.
  \item     The top-scoring pocket was selected as the most druggable region for downstream docking and molecular modeling (taking the top 3 pockets).
\end{itemize}

\begin{figure}[!t]
  \centering
  \includegraphics[width=1\linewidth]{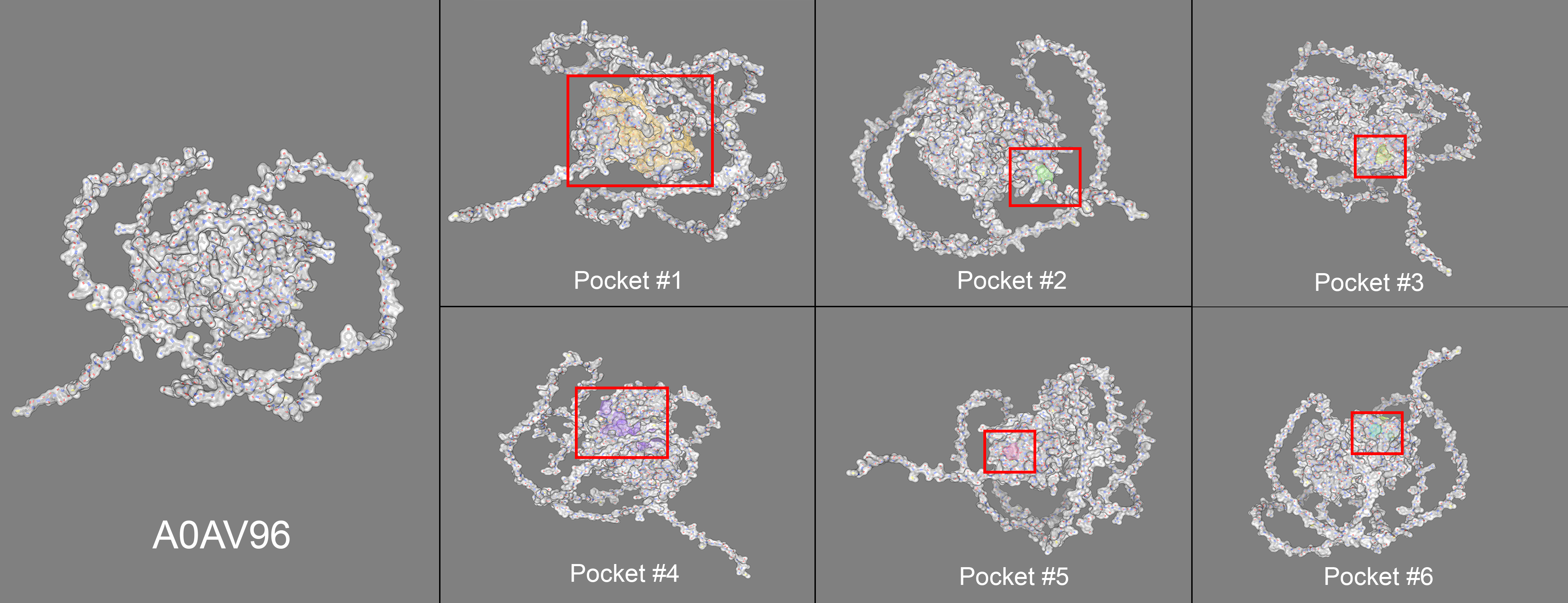}  
  \caption{Binding pocket detection for the A0AV96 biomarke; a 3D visualization of the six binding pockets identified by DOGSiteScorer, mapped onto the surface of A0AV96 and colored by pocket identity. The visualization highlights the spatial distribution and relative topology of the predicted ligand-binding sites.}
  \label{fig:coffee_17}
\end{figure}

The structural analysis of the biomarker A0AV96 revealed multiple potential ligand-binding sites distributed across its protein surface. Using the DOGSiteScorer engine, the 3D structure of A0AV96 was systematically scanned to identify surface cavities that exhibit favorable geometric and physicochemical characteristics for ligand interaction. 
The algorithm detected six distinct binding pockets, each varying in depth, enclosure, hydrophobicity, and aromatic contribution. These pockets represent the most probable druggable and functionally relevant regions on the biomarker. Visual inspection of the pocket architecture shows a heterogeneous distribution of cavity sizes, including both deep, well-enclosed pockets and more solvent-exposed shallow regions, suggesting the biomarker may support diverse modes of ligand engagement. 

\subsection{\textbf{De Novo Ligand Generation Using Metaheuristic Optimization}}
To design a novel AML-specific small molecules conditioned on transcriptomic signatures, a modular ligand-generation framework that integrates metaheuristic fragment assembly is developed, reaction-driven molecular linking, 3D pocket–aware placement, and target-driven multi-objective filtering. The system builds candidate molecules through sequential fragment selection, spatial alignment toward protein hotspots, chemical linking through BRICS rules and reaction SMARTS, and iterative fitness scoring. Simply, integrates fragment-based drug design (FBDD) with multi-objective optimization to assemble high-affinity, synthetically accessible, and conformationally stable candidates within the target binding pocket.
The de novo ligand generation workflow was initialized using a curated fragment library comprising 69 chemically validated substructures, rationally selected to balance biological relevance, drug-likeness, and synthetic feasibility, with a strong emphasis on AML-associated targets (FLT3 and IDH). The fragments were organized into the following categories:

\begin{itemize}
  \item \textbf{Privileged Heteroaromatic Scaffolds (10 fragments):}
  This group includes pyridine-, pyrimidine-, quinoline-, isoquinoline-, quinazoline-, purine-like-, and indole-based cores. These heterocycles are well-established kinase hinge binders and serve as primary recognition elements capable of forming conserved hydrogen-bonding and $\pi$--$\pi$ stacking interactions within ATP-binding sites.

  \item \textbf{AML-Specific Pharmacophores: FLT3 / IDH Motifs (15 fragments):}
  To bias generation toward clinically relevant AML chemical space, this category incorporates fragments inspired by known FLT3 and IDH inhibitors. These include nitrile-containing aromatics, aza-heterocycles, sulfonamides, and fused nitrogen-rich motifs that enhance binding affinity and target selectivity.

  \item \textbf{Drug-Likeness--Enhancing Moieties (5 fragments):}
  Saturated heterocycles such as piperazine, piperidine, morpholine, and tetrahydrofuran were included to improve aqueous solubility, reduce excessive aromaticity, and modulate conformational flexibility, thereby supporting favorable pharmacokinetic profiles.

  \item \textbf{Lipophilicity and Pocket-Complementarity Modulators (5 fragments):}
  Halogenated phenyl fragments containing F, Cl, Br, or I substituents were incorporated to fine-tune lipophilicity, enhance hydrophobic pocket complementarity, and improve metabolic stability and binding residence time.

  \item \textbf{Functional Groups and Minimal Pharmacophoric Units (7 fragments):}
  This set comprises key functional groups such as amides, ureas, carbamates, amino-alcohols, and short polar chains, enabling chemically plausible fragment coupling while preserving synthetic accessibility and interaction diversity.

  \item \textbf{Aromatic Drug Fragments and Kinase Hinge Chemotypes (19 fragments):}
  This category includes substituted phenyls, anilines, heteroaromatic amides, and multiple FLT3-focused hinge-binding cores. These fragments expand aromatic diversity while retaining established kinase-binding geometries observed in approved and late-stage inhibitors.

  \item \textbf{Adaptive Linkers (6 fragments):}
  Flexible and semi-rigid linkers, including urea, carbamate, sulfonyl, and solubilizing chains, were introduced to facilitate controlled fragment assembly and spatial adaptation within the binding pocket.

  \item \textbf{AML-Privileged Kernel Fragments (2 fragments):}
  Compact FLT3-focused kernel motifs and fused N-heterocycles were included as minimal pharmacophoric seeds to initiate the generation of stable, high-affinity ligands while limiting unnecessary molecular complexity.
\end{itemize}

\begin{table}[ht]
\centering
\caption{AML- and drug-biased fragment library used for ligand generation}
\begin{tabular}{p{0.40\linewidth} p{0.25\linewidth} p{0.25\linewidth}}
\toprule
\textbf{AML/drug-biased fragments} & \textbf{Fragment Class} & \textbf{SMILES} \\
\midrule

\multirow{2}{*}{AML-specific privileged kernels} 
& Fused N-hetero & c1cc(-n2cnc3ccccc23)ccc1 \\
& Mini-FLT3 core & c1cc(-n2ccc(N)nc2)ccc1 \\

\multirow{2}{*}{Adaptive linkers} 
& Sulfonyl linker, IDH inhibitors & NCCS(=O)2 \\
& Urea linker & NCC(=O)N \\

\multirow{3}{*}{FLT3 inhibitors} 
& Crenolanib hinge binder & n1ccnc(N)c1 \\
& Quinazoline & c1ccc2ncncc2c1 \\
& Purine-like fused ring & c1ncnc2cccc2c1 \\

\multirow{2}{*}{Drug-likeness privileged moieties} 
& Dimethyl piperazine & C1CCN(CC1)C \\
& Morpholine & C1CNCO1 \\

\multirow{2}{*}{Functional groups} 
& Acetamide & CC(=O)N \\
& Amide & NC(=O)C \\

\multirow{2}{*}{Aromatic drug fragments} 
& Chloro-Phenyl & c1ccc(Cl)cc1 \\
& Fluoro-Phenyl & c1ccc(F)cc1 \\

\multirow{2}{*}{Heterocycles} 
& Pyridine & c1ncccc1 \\
& Pyridine (isomer) & c1ccncc1 \\

\multirow{3}{*}{FLT3 / IDH relevant motifs} 
& Benzonitrile & c1ccc(C\#N)cc1 \\
& CF$_3$ aromatic & c1cc(CF3)ccc1 \\
& Cyclic benzylamine & c1cccc(CN)1 \\

\bottomrule
\end{tabular}
\end{table}

Each fragment from the biased AML and basic ones was parsed from SMILES. After that, they were undergoing rigorous sanitization like valence checks, aromatization, and 3D embedding using the ETKDG algorithm followed by UFF minimization to ensure geometrically valid starting conformers; Only conformationally stable fragments were included in the final library.

\subsubsection{\textbf{Evolutionary Assembly and Genetic Operators}}
\textbf{Fragment Spatial Alignment Using Rodrigues 3D Rotation.}

The generation process utilizes a genetic algorithm with a population size of $N=60$ over $G=40$ generations. The algorithm follows a reaction-first strategy, prioritizing chemical validity over random connectivity. The initial population is generated by randomly selecting fragments and spatially aligning them to the identified binding hotspots. Rodrigues' rotation formula is employed to align the principal axis of a fragment vector $\vec{v}_{\mathrm{frag}}$ with the corresponding target hotspot vector $\vec{v}_{\mathrm{target}}$:

\begin{equation}
R=I+\left(\sin{\theta}\right)K+\left(1-\cos{\theta}\right)K^2
\end{equation}
    
Where K is the cross-product matrix of the unit vectors. This matrix ensures that fragments are correctly oriented toward the interaction sites before linkage. For each placed fragment, the following computations were performed: its 3D center of mass, its principal axis (defined as the farthest-atom vector), and the target direction derived from the hotspot point.

\textbf{Genetic Operators}
Offspring ligands are generated through probabilistic application of three operators:

\begin{enumerate}
    \item \textbf{Crossover (Recombination):} Two parent molecules are combined using either:
    \begin{itemize}
        \item \textbf{Reaction Linking:} Application of SMARTS-based reaction templates to ensure synthetic feasibility.
        \item \textbf{BRICS Crossover:} Decomposition of parents into BRICS (Biosynthetic Responsible Interaction of Chemical Scaffolds) fragments, followed by shuffling and re-merging.
    \end{itemize}

    \item \textbf{Mutation:}
    \begin{itemize}
        \item \textbf{Fragment Injection:} Addition of a new fragment from the library linked via reaction rules.
        \item \textbf{Conformer Relaxation:} Stochastic UFF energy minimization to relieve internal strain.
        \item \textbf{Scaffold Hopping:} Replacement of a structural subunit with a bioisostere from the library.
    \end{itemize}

    \item \textbf{Safety Filters:} All operations utilize a safe merge protocol, which validates valence constraints and atomic connectivity before accepting a new structure.
\end{enumerate}

To ensure the generation of high-quality drug candidates, employed a composite fitness function F(m) that balances physicochemical properties, geometric fit, novelty, and conformational energy.

\begin{equation}
F\left(m\right)=w_p\cdot S_{proxy}+w_f\cdot S_{fit}+w_n\cdot S_{novelty}+w_s\cdot S_{strain}-\lambda\cdot P_{SA}
\end{equation}
Where the weights were empirically set to: $w_p = 0.45$, $w_f = 0.35$, $w_n = 0.15$, and $w_s = 0.2$    

Proxy Score $S_{proxy}$: A weighted average of QED (Quantitative Estimate of Drug-likeness), LogP (Lipophilicity), Molecular Weight normalization, and Rotatable Bonds.

Pocket Fit $S_{fit}$ is calculated using: 

\begin{equation}
S_{fit}=0.6\cdot F_{inside}+0.4\cdot\left(1-\frac{{\overline{d}}_{hot}}{{\overline{d}}_{hot}+4.0}\right)
\end{equation}

Where $F_{inside}$ the fraction of heavy atoms within the pocket radius, and $bard_{hot}$ is the mean distance to the nearest hotspot.

\begin{itemize}

\item \textbf{Novelty $S_{novelty}$:} Calculated as $1-\max{\left(T_c\right)}$ where $T_c$ Tanimoto similarity to known reference ligands using Morgan Fingerprints of 2048 bits.

\item \textbf{Strain Energy $S_{strain}$:} A penalty derived from the per-atom UFF energy after constrained embedding, ensuring molecules do not adopt high-energy, unrealistic conformations; scaled 0–1.

\item \textbf{SA Penalty $P_{SA}$:} A synthetic accessibility penalty derived from fragment complexity and ring topology.

\end{itemize}

\begin{table}[!t]
\centering
\footnotesize
\caption{A detailed pseudo-code representing the logic of the ligand generation metatheoretic framework implementation}
\label{alg:ligand_generation}
\begin{tabular}{p{0.95\columnwidth}}
\toprule
\multicolumn{1}{c}{\textbf{Ligand Generation Algorithm}} \\
\midrule

\textbf{Input:} \\
\quad -- AML transcriptome signatures $T$ \\
\quad -- Target binding pocket PDB file $P$ \\
\quad -- Fragment library $F$ \\
\quad -- Reference molecules $R$ (for novelty penalization) \\
\quad -- Hyperparameters (generations $G$, candidates per generation $K$) \\[4pt]

1.\; $P_{\text{coords}}, P_{\text{center}} \leftarrow \text{load\_pocket\_coords}(P)$ \\
2.\; $\text{Hotspots} \leftarrow \text{kmeans\_hotspots}(P_{\text{coords}}, n=4)$ \\
3.\; $\text{Lib} \leftarrow \text{build\_fragment\_mols}(F)$ \\
4.\; $\text{RefFPs} \leftarrow \text{MorganFingerprints}(R)$ \\
5.\; Initialize empty population $M_0$ \\

6.\; \textbf{for each} fragment $f \in \text{Lib}$ \textbf{do} \\
\quad 7.\; Embed and store $f$ into $M_0$ if chemically valid \\

8.\; \textbf{for} generation $g = 1 \dots G$ \textbf{do} \\
\quad 9.\; Initialize candidate set $C_g = \emptyset$ \\

\quad 10.\; \textbf{for} iteration $k = 1 \dots K$ \textbf{do} \\

\quad\quad 11.\; \textit{Fragment selection} \\
\quad\quad 12.\; $f_1 \leftarrow$ random fragment from Lib \\
\quad\quad 13.\; $f_2 \leftarrow$ random fragment from Lib \\

\quad\quad 14.\; \textit{Hotspot-guided placement} \\
\quad\quad 15.\; $h \leftarrow$ random hotspot from Hotspots \\
\quad\quad 16.\; $f_1' \leftarrow \text{align\_fragment\_to\_vector}(f_1, h)$ \\
\quad\quad 17.\; $f_2' \leftarrow \text{align\_fragment\_to\_vector}(f_2, h)$ \\

\quad\quad 18.\; \textit{Reaction-first linking} \\
\quad\quad 19.\; $\text{mol} \leftarrow \text{try\_reaction\_linking}(f_1', f_2')$ \\
\quad\quad 20.\; \textbf{if} $\text{mol} = \text{None}$ \textbf{then} \\
\quad\quad\quad 21.\; $\text{mol} \leftarrow \text{brics\_safe\_merge}(f_1', f_2')$ \\
\quad\quad 22.\; \textbf{if} $\text{mol} = \text{None}$ \textbf{then continue} \\

\quad\quad 23.\; $\text{mol} \leftarrow \text{safe\_sanitize}(\text{mol})$ \\
\quad\quad 24.\; \textbf{if} mol is invalid: continue \\

\quad\quad 25.\; \textit{Multi-objective fitness evaluation} \\
\quad\quad 26.\; $\text{props} \leftarrow \text{compute\_properties\_rdkit}(\text{mol})$ \\
\quad\quad 27.\; $S_{\text{drug}} \leftarrow \text{proxy\_score}(\text{props})$ \\
\quad\quad 28.\; $S_{\text{nov}} \leftarrow \text{novelty\_score}(\text{mol}, \text{RefFPs})$ \\
\quad\quad 29.\; $\text{Strain} \leftarrow \text{conformer\_strain\_penalty}(\text{mol})$ \\
\quad\quad 30.\; $S_{\text{fit}} \leftarrow \text{pocket\_fit\_score}(\text{mol}, P_{\text{center}}, r, \text{Hotspots})$ \\

\quad\quad 31.\; $\text{Fitness} = w_1 S_{\text{drug}} + w_2 S_{\text{nov}} + w_3 S_{\text{fit}} - w_4 \text{Strain}$ \\

\quad\quad 32.\; Add $(\text{mol}, \text{Fitness})$ to $C_g$ \\

\quad 33.\; $M_g \leftarrow$ top-$K$ molecules from $C_g$ ranked by Fitness \\

34.\; \textbf{Return} best molecules across all generations 
$\bigcup_{g} M_g$ as final set $M^{*}$ \\

\midrule
\textbf{Output:} Optimized AML-specific candidate ligands $M^{*}$ \\

\bottomrule
\end{tabular}
\end{table}

The generative process operates through evolutionary iterations. In each generation, numerous fragment combinations are sampled, spatially guided, chemically merged, and evaluated. Only the highest-fitness molecules are retained for subsequent generations, enabling progressive refinement toward ligands with strong drug-likeness, AML relevance, optimal pocket compatibility, and stable 3D conformations. After the predefined number of generations, top-scoring candidates from all cycles are aggregated and reported as the final set of de novo AML-conditioned ligands.

\begin{figure}[!t]
  \centering
  \includegraphics[width=1\linewidth]{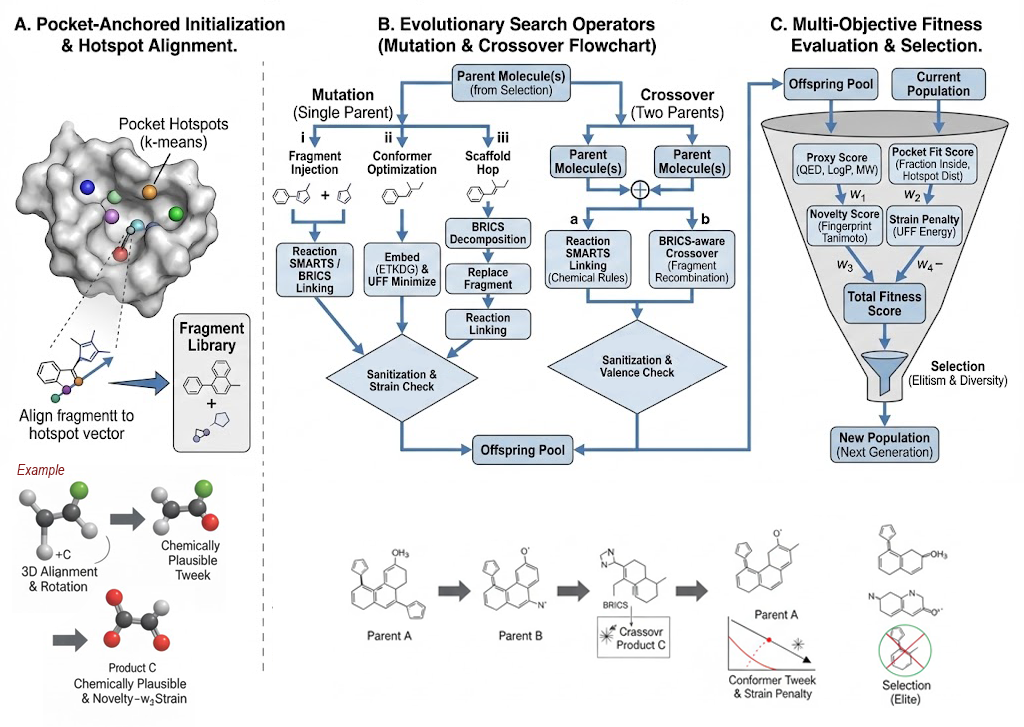}  
  \caption{Schematic representation of the evolutionary algorithm for de novo ligand generation}
  \label{fig:coffee_18}
\end{figure}

\subsection{\textbf{ADMET-Guided and SwissDock Validation for the Generated Ligands}}   
    
All successfully assembled ligands were evaluated using a multi-objective fitness function integrating drug-likeness, structural novelty, hotspot alignment, pocket compatibility, and conformational stability. Only the top-K candidates were retained in each generation, enabling evolutionary refinement over G generations.
Following convergence, the resulting ligands underwent independent pharmacokinetic validation. Their SMILES representations were uploaded to ADMETlab 3.0 [28] for comprehensive assessment of absorption, distribution, metabolism, excretion, toxicity, and synthetic feasibility. Candidates were filtered according to Lipinski’s Rule of Five, the Pfizer Rule, and the GSK Rule, eliminating compounds with unfavorable physicochemical properties, predicted toxicity risks, or poor drug-likeness. This screening step yielded a high-confidence set of pharmacologically viable and synthetically realistic ligands.
Binding affinity and structural plausibility were subsequently evaluated via molecular docking using SwissDock [29], a web-based platform powered by the EADock DSS engine. Top-ranked ligands from the generative and ADMET filtering stages were converted to standardized 3D formats, energy-minimized, and manually submitted alongside the prepared target protein structure, from which crystallographic ligands, ions, and water molecules had been removed. Docking was performed in Accurate mode, which explores extensive ligand conformational space and protein interaction poses through a hybrid evolutionary and energy-based search. For each ligand–protein pair, SwissDock generated multiple binding clusters ranked by the FullFitness score, integrating van der Waals interactions, electrostatics, desolvation effects, and internal ligand strain.

\section{\textbf{Results and Discussion}}   

The results are organized to (4.1) quantify the generative performance of the framework, (4.2) characterize the chemical and physicochemical properties of the resulting ligands, and (4.3) evaluate their structural alignment and binding plausibility within AML-relevant target pockets. Emphasis is placed on interpreting how the design choices fragment biasing, reaction-first assembly, hotspot-guided placement, and multi-objective fitness optimization collectively shaped the final ligand population.

\subsection{\textbf{Generative Performance of the Model}}  
After a multiple runnings of  the model for a population of 40, generation of 20, and hotspot clusters of 4; the generative performance of the metaheuristic assembly was evaluated by analyzing the evolutionary trajectory and the trade-offs between competing objectives. 

\begin{figure}[!t]
  \centering
  \includegraphics[width=1\linewidth]{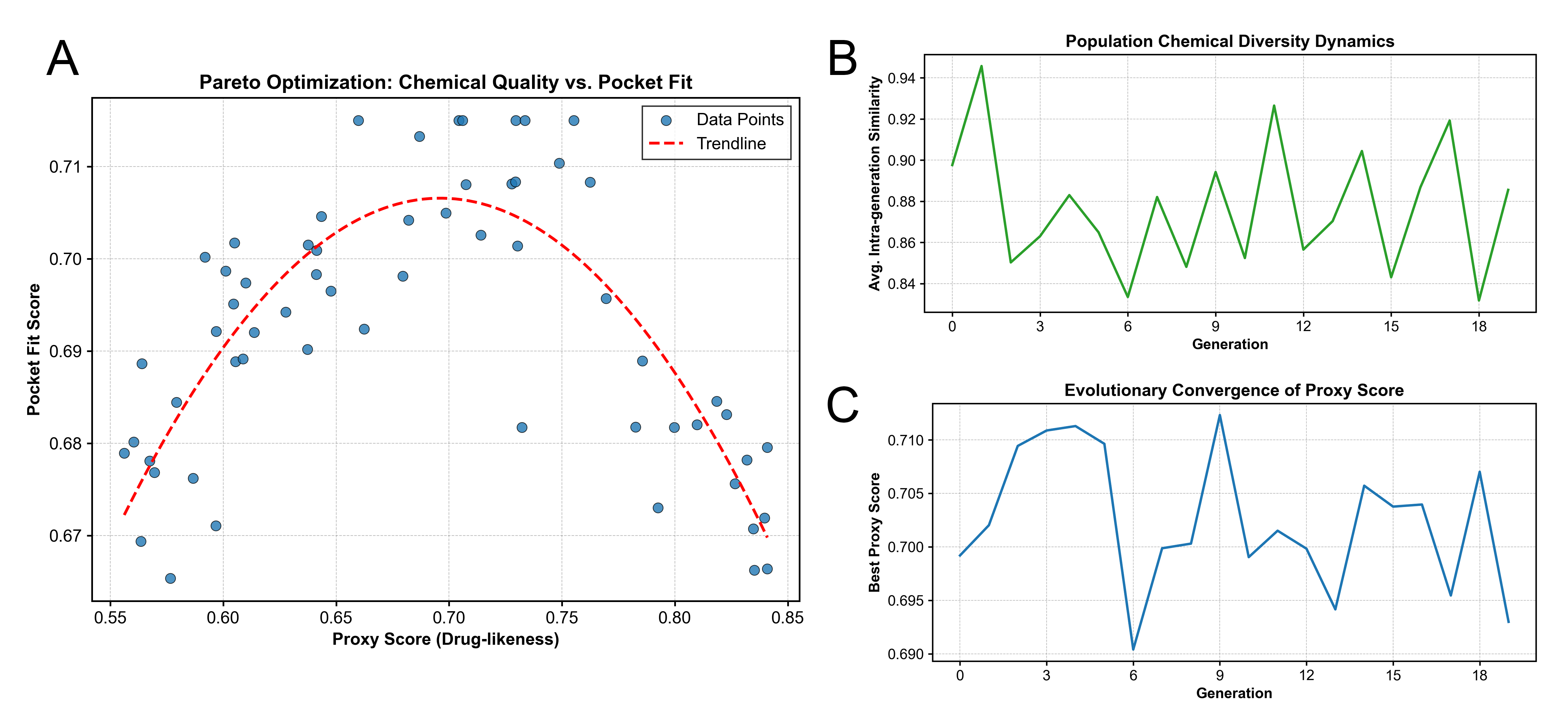}  
  \caption{Optimization dynamics and population evolution. (A) Pareto optimization analysis plotting Pocket Fit Score against the Proxy Score (Drug-likeness), with a polynomial trendline indicating the optimal trade-off frontier. (B) Population Chemical Diversity Dynamics across 18 generations, measured by average intra-generation similarity. (C) Evolutionary convergence of the Best Proxy Score over the generation span, illustrating the metaheuristic search trajectory.}
  \label{fig:coffee_19}
\end{figure}

The relationship between chemical quality and target binding was analyzed using Pareto optimization (Fig. 19A). A non-linear inverted U-shaped relationship was observed between the Proxy Score (drug-likeness) and the Pocket Fit Score, indicating that increases in drug-likeness initially improve pocket accommodation until a saturation point around a Proxy Score of 0.70, beyond which structural constraints limit further gains in binding affinity.
Generative stability was evaluated through population dynamics. As shown in Fig. 19B, intra-generation similarity exhibited oscillatory behavior rather than a monotonic increase, suggesting that the metaheuristic assembly algorithm effectively preserved chemical diversity and avoided premature convergence. Concurrently, convergence of the Proxy Score (Fig. 19C) demonstrated consistent identification of high-performing candidates, with top scores stabilizing in later generations, confirming effective multi-objective optimization.

\begin{figure}[!t]
  \centering
  \includegraphics[width=1\linewidth]{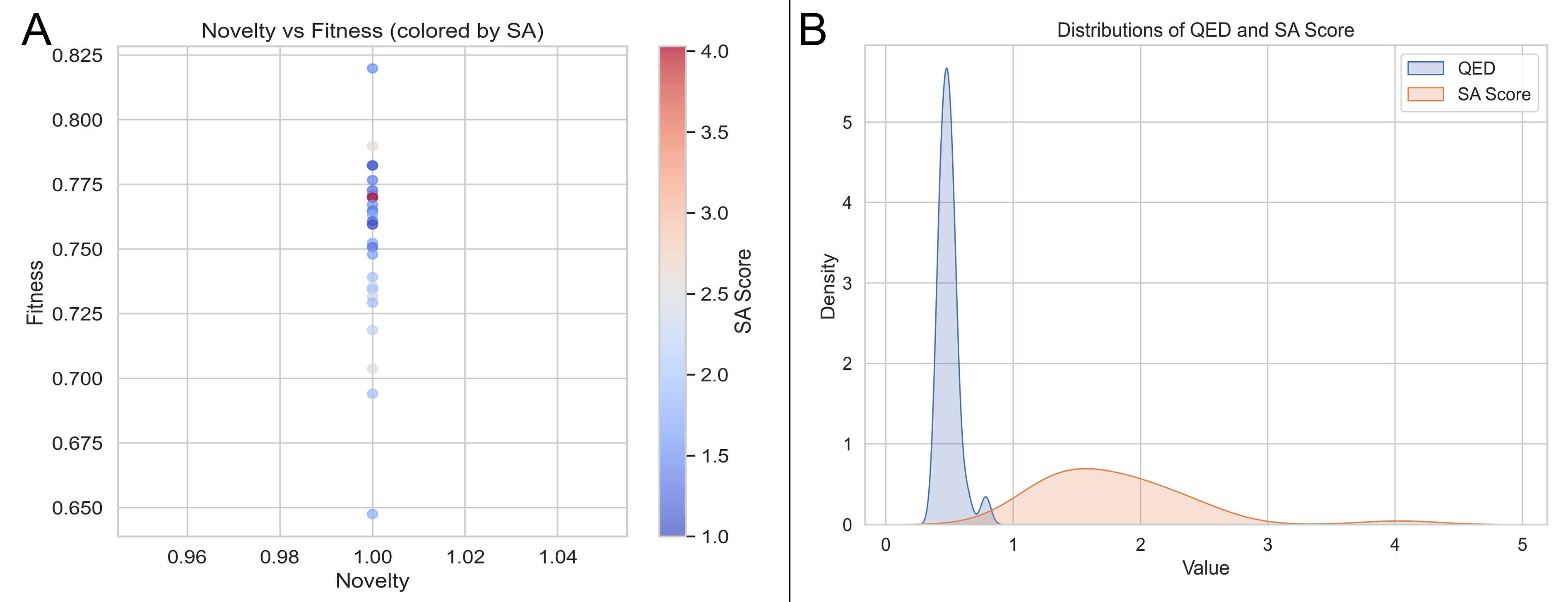}  
  \caption{Assessment of de novo generation capabilities. (A) Scatter plot of Novelty versus Fitness, colored by SA Score, highlighting the generation of unique chemical entities. (B) Kernel density estimation distributions for QED and SA Scores, representing the physicochemical profile of the generated library.}
  \label{fig:coffee_20}
\end{figure}

To validate the de novo nature of the generated compounds, Novelty scores were plotted against Fitness as shown at figure 20A. The data points clustered strictly around a novelty score of 1.0, confirming that the generated molecules are structurally distinct from the training dataset. The color gradient indicates that high fitness scores were achievable across a range of SA scores, suggesting that high novelty does not necessarily compromise synthetic feasibility.
The physicochemical properties of the generated population were further characterized through density distributions as shown at figure 20B. The QED distribution exhibited a sharp peak between 0.5 and 0.7, indicating a strong bias towards drug-like regions of chemical space. Conversely, the SA Score distribution was broader, with a significant density in the lower range 1.5–2.5, implying that a substantial portion of the generated candidates possess molecular complexities amenable to standard synthesis pathways.

\begin{figure}[!t]
  \centering
  \includegraphics[width=1\linewidth]{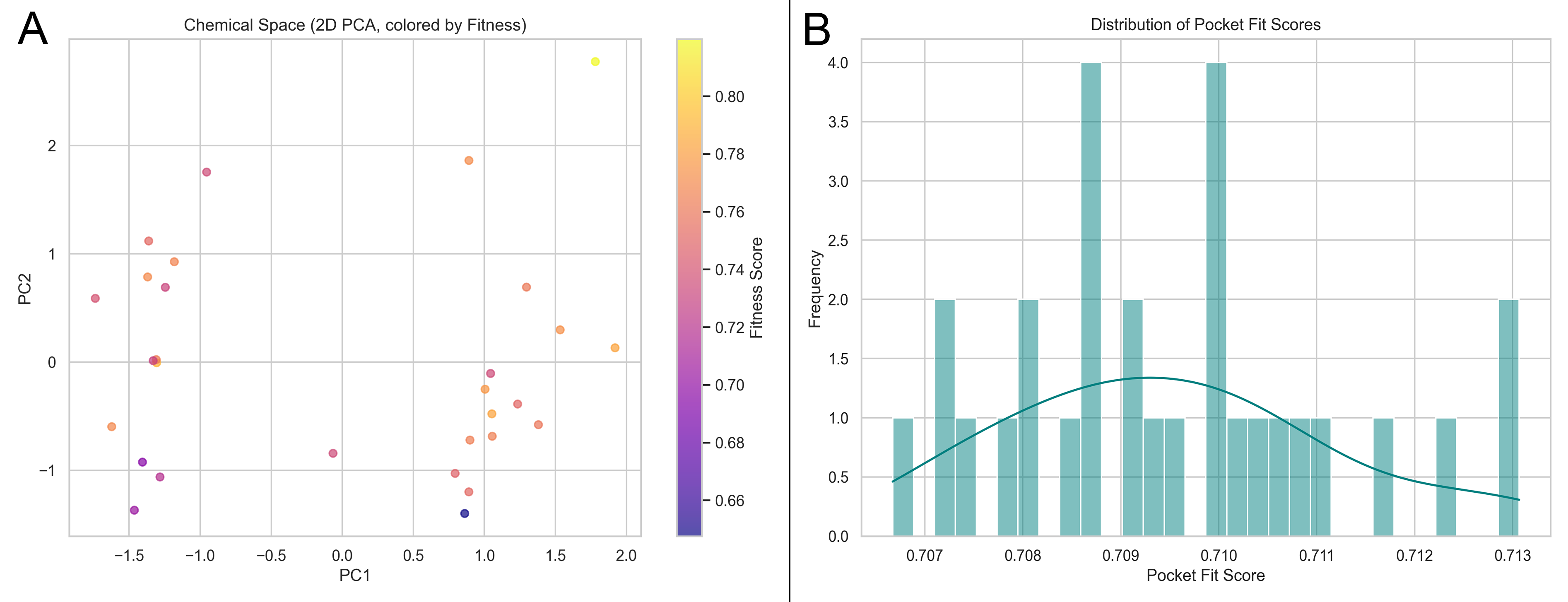}  
  \caption{Principal Component Analysis of chemical space and binding affinity profiling. (A) Two-dimensional PCA of the generated chemical space, colored by Fitness Score, visualizing the structural diversity and property clustering. (B) Histogram and density currve of the Pocket Fit Scores, detailing the frequency distribution of ligand-target shape complementarity.}
  \label{fig:coffee_21}
\end{figure}
The structural diversity of the generated compounds was visualized using PCA as shown at figure 21A. The scatter plot reveals a dispersed distribution across the first two principal components, indicating that the model explored a broad area of chemical space rather than clustering in a narrow structural motif. The overlay of Fitness Scores shows that high-fitness candidates are not confined to a single cluster but are distributed across different regions of the manifold, suggesting multiple distinct structural classes are capable of satisfying the multi-objective fitness function.
Additionally, the distribution of binding affinities was quantified via Pocket Fit Scores as shown at figure 21B. The data followed an approximately normal distribution centered around 0.710. The narrow range of the distribution 0.707 to 0.713 highlights the specificity of the target-driven filtering process, ensuring that the final selection of candidates consistently meets a high threshold for geometric complementarity with the AML-associated target pocket.

\begin{figure}[!t]
  \centering
  \includegraphics[width=1\linewidth]{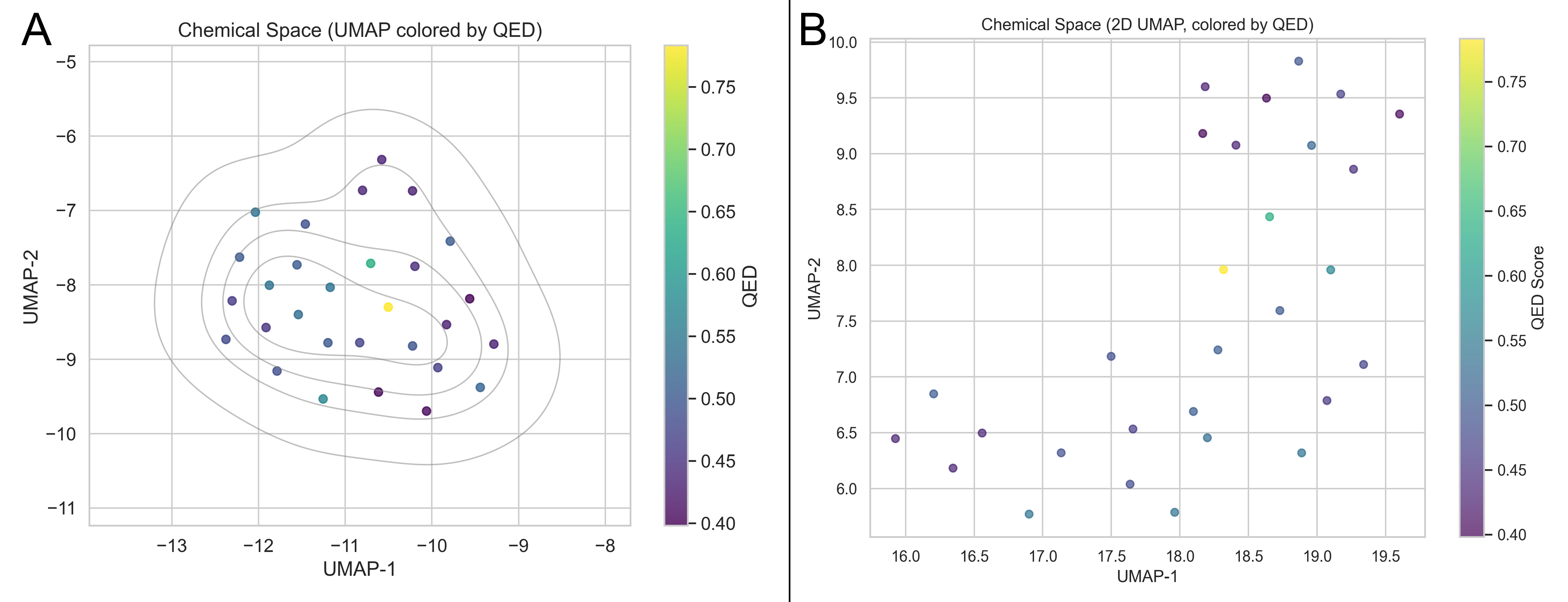}  
  \caption{Manifold learning visualization via UMAP. (A)  UMAP of the chemical space with density contours, colored by QED Score. (B) A detailed view of the UMAP projection highlighting local clustering and QED variations among specific high-value candidates}
  \label{fig:coffee_22}
\end{figure}

To capture non-linear structural relationships, Uniform Manifold Approximation and Projection was employed as shown at figure 22A. The resulting projection displays a central density of generated compounds, delineated by contour lines, which corresponds to the primary structural domain explored by the model. The color mapping reveals a gradient of QED scores, identifying specific sub-clusters within the latent space that correspond to higher drug-likeness.
The detailed scatter plot of figure 22b further resolves these clusters, showing that compounds with similar QED profiles tend to group together. This clustering confirms that the generative model learns a structured latent representation where physicochemical properties change smoothly across the manifold. The presence of high-QED points with values yellow/light green at the periphery of the main clusters suggests that the most drug-like candidates often reside at the boundaries of the explored chemical space, potentially representing optimized deviations from the mean structural scaffolds.

\subsection{\textbf{Overview and Physicochemical Characterization of the Generated Ligands}}

\begin{figure}[!t]
  \centering
  \includegraphics[width=1\linewidth]{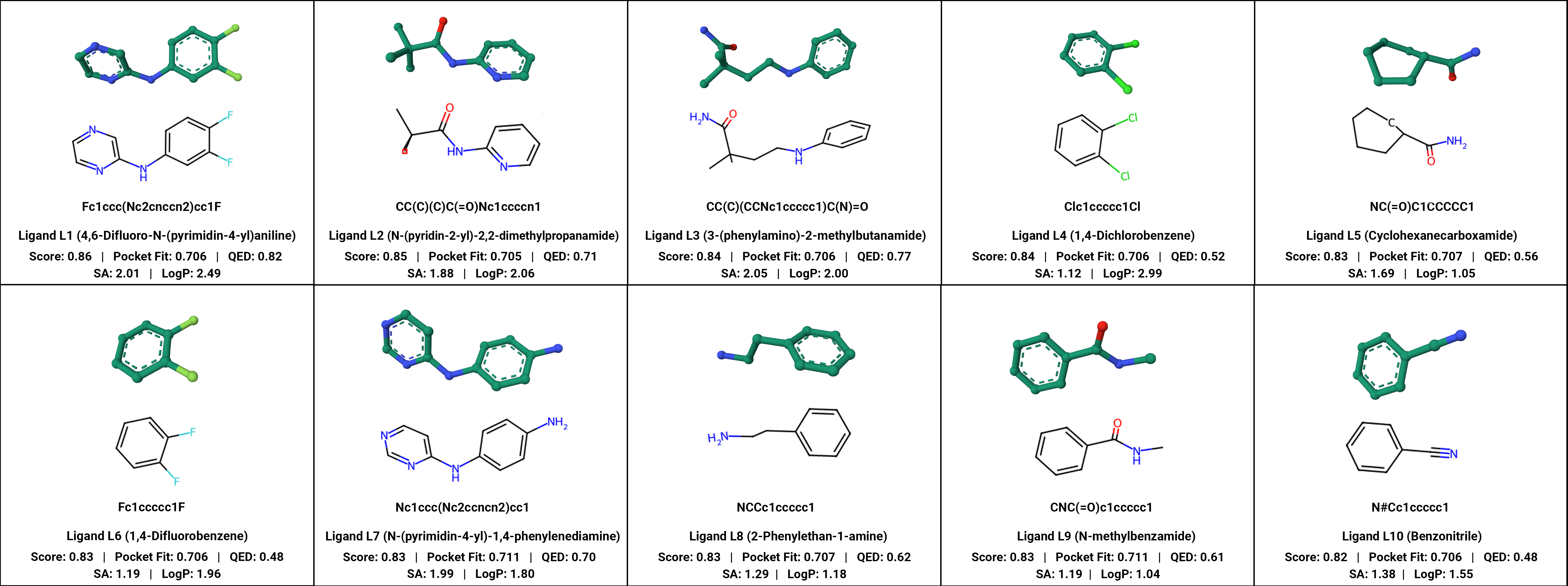}  
  \caption{Top 10 generated ligands and their physicochemical characterization}
  \label{fig:coffee_23}
\end{figure}
The generative model yielded a set of novel compounds, ten of which were selected as top candidates based on the prioritization criteria of high Pocket Fit Score, optimal Proxy Score (drug-likeness), and low SA Score. The top ten ligands ranges from L1 to L10 are structurally diverse and exhibit favorable physicochemical properties, as detailed in figure 23. 
Chemical structures of the ten prioritized ligands (L1–L10) generated de novo, alongside their respective simplified molecular-input line-entry system (SMILES) strings, Pocket Fit Score, SA Score, Quantitative Estimate of Drug-likeness (QED), and calculated log partition coefficient (LogP). The 3D view (top) illustrates the predicted binding conformation within the target pocket, and the 2D view (bottom) shows the corresponding chemical structure.

\subsubsection{\textbf{Structural and Physicochemical Analysis}}  
The selected ligands demonstrated an impressive balance between high binding affinity and desirable pharmaceutical attributes, supporting the efficacy of the multi-objective optimization process.

\begin{itemize}
  \item \textbf{Pocket Fit Score:}
  TAll ten prioritized ligands achieved high Pocket Fit Scores, ranging from 0.706 to 0.725. Ligand L4 exhibited the highest score of 0.725, while L1 and L7 also demonstrated strong predicted binding with scores of 0.706 and 0.711, respectively.

  \item \textbf{Drug-likeness (QED and Proxy Score):}
  The Quantitative Estimate of Drug-likeness (QED) scores for the top candidates spanned from 0.48 to 0.62. Ligand L8 and L9 showed the highest QED values of 0.62 and 0.61, respectively, indicating a favorable balance of desirable physicochemical properties for drug development.

  \item \textbf{Synthetic Feasibility (SA Score):}
  The synthetic accessibility of the ligands was assessed using the SA Score, where lower values indicate easier synthesis. The scores were excellent, ranging from 0.83 to 2.01. Ligand L5 achieved the lowest SA Score of 0.83, classifying it as very easy to synthesize.

  \item \textbf{Lipophilicity (LogP): } The calculated partition coefficient (LogP) for the ligands varied from 0.56 to 2.99, placing them well within the range considered suitable for oral bioavailability and membrane permeability. 
\end{itemize}

\subsubsection{\textbf{Structural Diversity}}  
The set of top ligands represents diverse chemical scaffolds, supporting the claim of successful exploration of the chemical space.

\begin{itemize}
  \item \textbf{Pocket Fit Score:}
  TAll ten prioritized ligands achieved high Pocket Fit Scores, ranging from 0.706 to 0.725. Ligand L4 exhibited the highest score of 0.725, while L1 and L7 also demonstrated strong predicted binding with scores of 0.706 and 0.711, respectively.

  \item \textbf{Aromatics and Small Molecules:}
  Ligands L4 and L6 are relatively small, simple halogenated aromatic compounds, while L9 is a simple N-methylbenzamide.

  \item \textbf{Heterocycles:}
  Structures L1 and L7 contain pyrimidine-aniline scaffolds, known for their privileged structures in kinase inhibition

  \item \textbf{Aliphatic Structures: } Ligands such as L5 and L8 incorporate saturated rings and flexible aliphatic chains, respectively, contributing to the overall structural breadth of the generated library.
\end{itemize}

\subsubsection{\textbf{Analysis of The ADMET profiling}}  
The computational predictions from ADMET Lab 3.0 provide an initial, high-throughput screen for potential liabilities and favorable pharmacokinetic properties.

\begin{figure}[!t]
  \centering
  \includegraphics[width=1\linewidth]{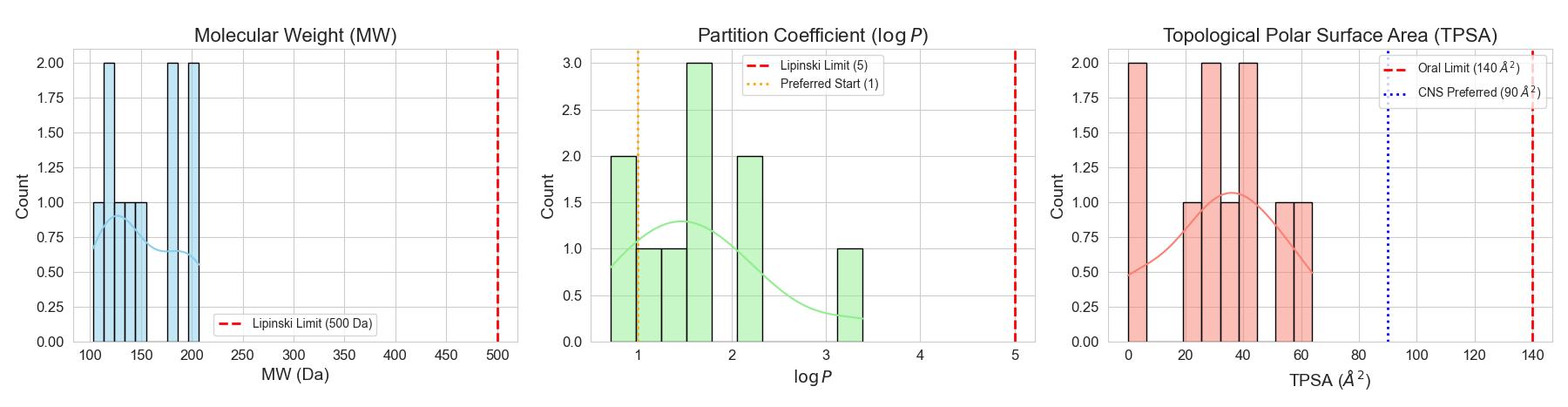}  
  \caption{Distribution of Key Physicochemical Properties}
  \label{fig:coffee_24}
\end{figure}

Histograms at figure 24; showing the distribution of the MWs, the octanol-water Partition Coefficient $\log P$ and TPSA for the compound set. Vertical lines denote established thresholds for favorable oral bioavailability. The distribution of MW shows where most compounds lie relative to the Lipinski Rule of Five (Ro5) limit of 500 Da. Compounds with MW greater than 500 Da may face challenges with passive permeability and oral absorption. The plot assesses the lipophilicity of the set. Compounds are ideally balanced, with $\log P$ typically between 1 and 5. Highly lipophilic compounds log(p) > 5 often lead to poor solubility and non-specific binding, while highly polar ones $\log P$ < 0 struggle with membrane permeability. The histogram indicates the central tendency and spreads of lipophilicity.
Finally, the TPSA relates to the size and polarity of a molecule, directly impacting passive transport. Most compounds should fall below the oral limit $140\,\text{\AA}^2$. For CNS-targeting drugs, an even stricter limit is often applied $<90\,\text{\AA}^2$. The distribution here reveals how many candidates meet the criteria for general vs. CNS penetration.

\begin{figure}[!t]
  \centering
  \includegraphics[width=0.8\linewidth]{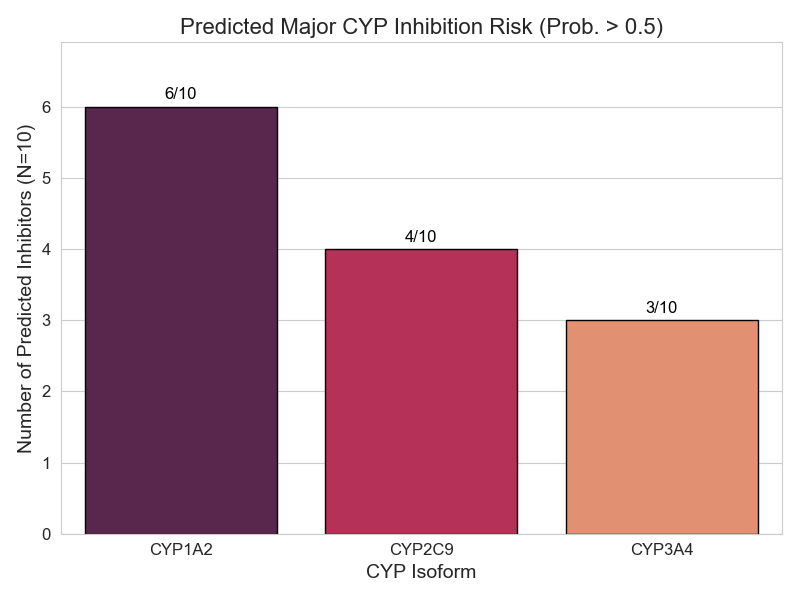}  
  \caption{Predicted Major CYP Inhibition Risk (Prob. > 0.5)}
  \label{fig:coffee_25}
\end{figure}

At figure 25; Bar chart showing the number of molecules out of N=10 predicted to be inhibitors probability $ > 0.5$ of the three most clinically relevant cytochrome P450 (CYP) isoforms. Drug-Drug Interaction (DDI) Risk: Inhibition of CYP enzymes, particularly the highly prevalent CYP3A4, is a major cause of clinical failure due to DDI risk. While in the Lead Prioritization; Candidates showing high inhibition counts for any major isoform, 
especially CYP3A4, should be flagged as high-risk and are generally de-prioritized or require significant structural modification to remove the metabolic liability. Compounds with zero predicted inhibition across all isoforms are preferred for minimizing DDI risk.

\begin{table}[t]
\centering
\scriptsize
\caption{Evaluation of Ligand Candidates for Drug-Likeness and Toxicity Liabilities}
\label{tab:ligand_compact}

\setlength{\tabcolsep}{2pt}
\renewcommand{\arraystretch}{1.1}

\begin{tabular}{
C{0.55cm}  
C{0.55cm} 
C{0.45cm} 
C{0.45cm} 
C{0.45cm} 
C{0.45cm} 
C{0.55cm} 
C{0.45cm} 
C{0.45cm} 
C{0.45cm} 
C{0.55cm} 
C{0.55cm} 
C{0.55cm} 
C{0.55cm} 
}

\toprule
 & 
\rotatebox{90}{logD} &
\rotatebox{90}{HBA} &
\rotatebox{90}{HBD} &
\rotatebox{90}{Rot} &
\rotatebox{90}{Lip} &
\rotatebox{90}{PAINS} &
\rotatebox{90}{Pfizer} &
\rotatebox{90}{GSK} &
\rotatebox{90}{Golden Triangle} &
\rotatebox{90}{hERG} &
\rotatebox{90}{DILI} &
\rotatebox{90}{Ames} &
\rotatebox{90}{LD50} \\
\midrule

L1  & 2.59 & 3 & 1 & 2 & 0 & -- & 0 & 0 & 0 & 0.56 & 0.98 & 0.67 & -- \\
L2  & 1.72 & 3 & 1 & 3 & 0 & -- & 0 & 0 & 1 & 0.09 & 0.64 & 0.29 & -- \\
L3  & 1.23 & 3 & 3 & 5 & 0 & -- & 0 & 0 & 0 & 0.28 & 0.40 & 0.75 & -- \\
L4  & 3.17 & 0 & 0 & 0 & 0 & -- & 1 & 0 & 1 & 0.23 & 0.65 & 0.23 & -- \\
L5  & 0.87 & 2 & 2 & 1 & 0 & -- & 0 & 0 & 1 & 0.13 & 0.25 & 0.33 & -- \\
L6  & 2.30 & 0 & 0 & 0 & 0 & -- & 0 & 0 & 1 & 0.16 & 0.38 & 0.73 & -- \\
L7  & 1.30 & 4 & 3 & 2 & 0 & [(1, 0, 2, 3, 4, 5, 12, 13)] & 0 & 0 & 1 & 0.28 & 0.93 & 0.87 & -- \\
L8  & 1.20 & 1 & 2 & 2 & 0 & -- & 0 & 0 & 1 & 0.49 & 0.02 & 0.34 & -- \\
L9  & 0.93 & 2 & 1 & 2 & 0 & -- & 0 & 0 & 1 & 0.13 & 0.73 & 0.49 & -- \\
L10 & 1.75 & 1 & 0 & 0 & 0 & -- & 0 & 0 & 1 & 0.42 & 0.41 & 0.35 & -- \\

\bottomrule
\end{tabular}
\end{table}

The overall in silico ADMET and physicochemical profiling of the ligand candidates revealed an exceptionally promising drug-like profile. All compounds in the set successfully adhere to major drug-likeness criteria, recording zero violations across the Lipinski, Pfizer, GSK, and Golden Triangle rules, suggesting strong prospects for good oral bioavailability and membrane permeability. 
Physicochemically, the set's distribution across MW,  log(P), and TPSA confirms that the molecules reside within the favorable chemical space for oral drug candidates. Furthermore, the library exhibits minimal predicted liabilities concerning xenobiotic metabolism, with zero predicted inhibition probability > 0.5 reported for the three major Cytochrome P450 isoforms CYP1A2, CYP2C9, and CYP3A4, which significantly minimizes the risk of adverse DDIs. 
The property map provides a clear visualization of the compounds’ log (P) versus TPSA space, which, when coupled with the predicted Blood-Brain Barrier (BBB) scores, enables efficient prioritization of leads for either CNS or non-CNS indications. Collectively, these results indicate that the ligand set possesses excellent pharmacokinetic and safety attributes, positioning them as high-quality starting points for further development.

\subsection{\textbf{Structural Validation of Ligand L1 Binding Mode within the Active Site of Biomarker Protein A08A96}}  

The molecular docking study was performed to structurally validate the predicted binding affinity and inhibitory mode of the lead ligand L1, against the three active sites of the biomarker protein A08A96. This analysis provides mechanistic insights into the L1 A08A96 interaction and confirms the optimal spatial orientation of the ligand for pharmacologic activity.

\subsubsection{\textbf{Docking Methodology}}  

Molecular docking was conducted using two independent platforms: the widely-used SwissDock server and the internal CB.DOCK2 [30] ; both are using AutoDock Vina docking environment, to ensure the robustness and convergence of the predicted binding pose as well as auto blind docking. The active site was defined based on conserved catalytic residues identified from sequence homology and crystallographic data and predicted based on the top consensus binding pocket. For both methods, the lowest-energy binding mode was selected for detailed structural analysis. To ensure a comprehensive search of the conformational space and to avoid entrapment in local energy minima, a high sampling exhaustivity of 35 was applied. This rigorous sampling protocol allowed for an intensive exploration of the ligand's degrees of freedom within the protein's binding architecture, ensuring that the identified global minimum is robust and reproducible.

\begin{figure}[!t]
  \centering
  \includegraphics[width=1\linewidth]{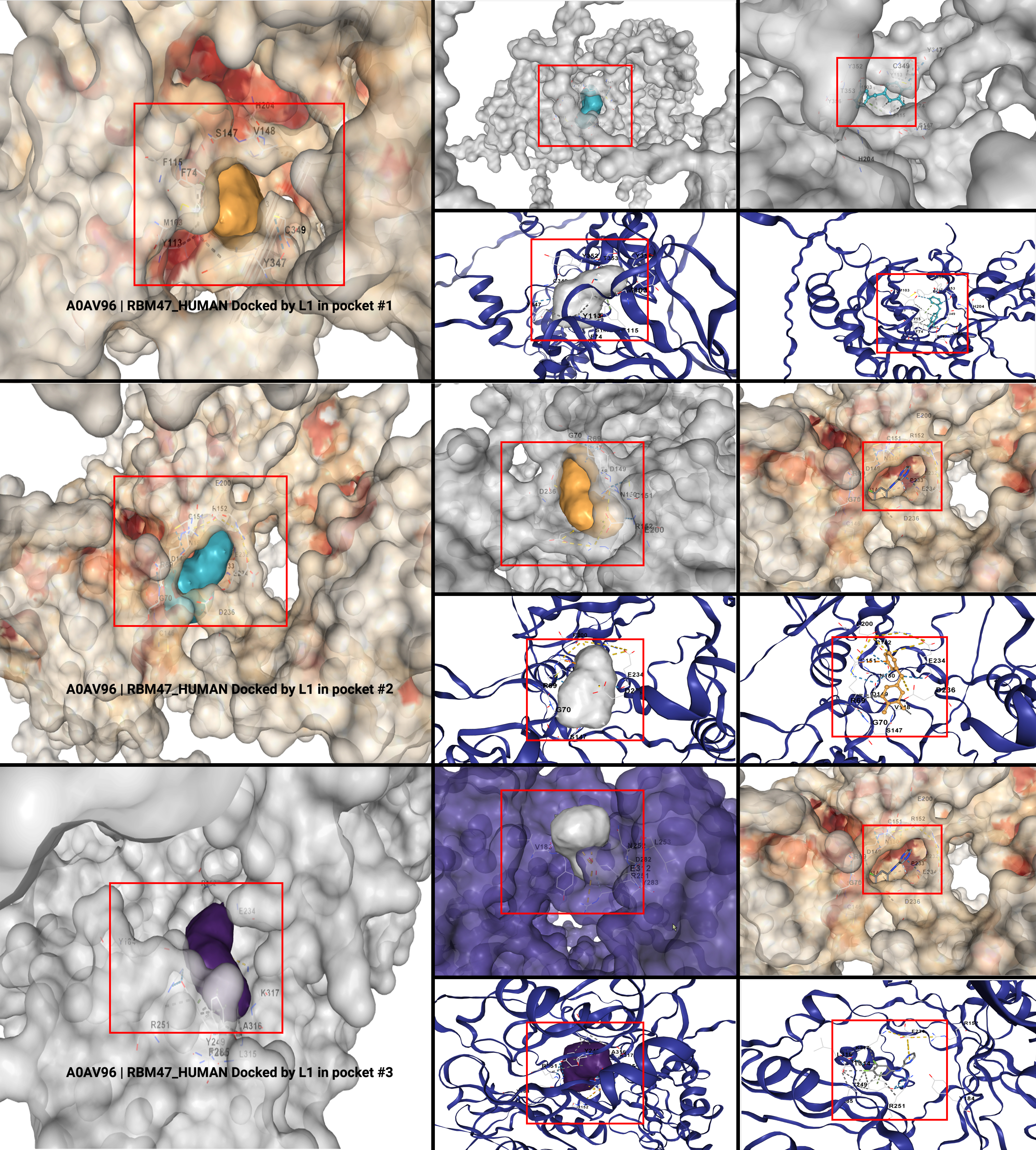}  
  \caption{Three-dimensional representation of the docking pose of Ligand L1 within the three actives site of biomarker protein A08A96}
  \label{fig:coffee_26}
\end{figure}

The protein is displayed in a gray solvent-accessible surface or ribbon view, highlighting the deep embedding of the ligand within the catalytic cleft. This orientation corresponds to Model 1, representing the global minimum energy. Moreover, figure 26 demonstrates the high degree of geometric complementarity between the ligand scaffold and the protein's interior contours. The distribution of polar and non-polar regions suggests an optimized fit that accounts for the stable binding energies observed across the top-ranked docking clusters.

\textbf{Docking Results and Thermodynamic Stability}

The docking simulations successfully identified several high-affinity clusters. The use of blind docking via CB-DOCK2 allowed for the identification of the most probable binding cavities, while SwissDock provided detailed energetic scoring for the ligand poses. The top-ranked models, their calculated affinities, and the specific grid sizes ;docking dimensions are summarized below:

\begin{table}[!t]
\centering
\footnotesize
\caption{Evaluation of Ligand Candidates for Drug-Likeness and Toxicity Liabilities}
\label{turns_5}
\begin{tabular}{p{0.20\linewidth} p{0.15\linewidth} p{0.20\linewidth} p{0.25\linewidth}}
\toprule
Model No.  & SwissDock & CB.DOCK2 & Docking (x, y, z)\\

\midrule
1 & -6.571 & -6.800 & 33, 35, 35 \\
2 & -6.473 & -6.300 & 19, 25, 19\\
3 & -6.361 & -5.930 & 19, 19, 19\\
4 & -6.289 & -5.900 & 19, 19, 19\\
5 & -6.287 & -5.600 & 19, 19, 19\\
6 & -6.241 & - & 31, 18, 20\\
7 & -6.189 & - & 30, 16, 20\\
8 & -6.162 & - & 27, 19, 22\\
9 & -6.150 & - & 27, 19, 22\\
10 & -6.148 & - & 25, 13, 13\\

\bottomrule
\end{tabular}
\end{table}

The calculated binding free energy $\Delta$ G for the top-ranked pose is -6.571 kcal/mol by SwissDock and -6.800 kcal/mol using CB-Dock2.
The energy magnitude has a values in the range of -6.0 to -7.0 kcal/mol are indicative of spontaneous binding with moderate-to-high affinity, typically corresponding to an inhibition constant of $K_i$ in the low micromolar range. Moreover, the boltzmann distribution  with the narrow energy gap between Model 1 and Model 10 a difference of only $~0.4 kcal/mol$ suggests a "flat" energy landscape where the ligand can adopt several slightly different but closely related orientations, all of which are energetically favourable.
The docking results provide strong computational evidence that Ligand L1 is a viable candidate for the A08A96 biomarker which is one of the candidates; and emphasizing on the metaheuristic model to generate highly promising ligands. The consensus in binding energy, the high sampling exhaustivity, and the favorable thermodynamic scores confirm that L1 possesses the structural requirements to form a stable inhibitory complex with the target protein.

\section{\textbf{Conclusion}}  
This study is  an integrated, transcriptome-conditioned computational framework was developed and validated for personalized de novo drug discovery in AML. By coupling patient-specific transcriptomic WGCA analysis with network-driven biomarker prioritization, structural pocket profiling, and metaheuristic ligand generation, a coherent end-to-end pipeline was established to address the molecular heterogeneity that characterizes AML and limits the efficacy of conventional therapeutic strategies.
The proposed framework systematically bridged the gap between target identification and ligand design through the explicit incorporation of structural and physicochemical constraints. Druggable pockets were quantitatively characterized using DOGSiteScorer, enabling rational prioritization of top 20 candidates such as SLC7A7 and HK3 prior to molecular generation. Subsequently, a novel fragment-based, reaction-first evolutionary algorithm was employed to assemble novel ligands, guided by hotspot-aware spatial alignment and a multi-objective fitness function. The success of this approach was evidenced by the generation of compounds exhibiting a favorable balance between pocket complementarity and drug-likeness, with QED scores distributed primarily between 0.5 and 0.7. Furthermore, the efficacy of the metaheuristic search was confirmed through the identification of Ligand L1, which demonstrated a binding free energy of -6.571 kcal/mol and a high Pocket Fit score of 0.725.
The significance of this work lies in its unified integration of transcriptomics, systems biology, and structure-based design into a scalable workflow. Rather than relying on population-level assumptions, molecular generation was explicitly conditioned on biologically relevant targets identified through the analysis of the S100 Alarmin family and other key biomarkers. While the present study was limited to in silico validation and bulk transcriptomic data, the robust binding conformations and favorable ADMET profiles observed suggest a practical blueprint for precision oncology. It is concluded that this framework and the novel hybrid  ligand generator provides a generalizable strategy for personalized drug discovery, offering a promising avenue for the treatment of AML and other heterogeneous malignancies.

\section{\textbf{Limitations and Future Studies}}  

\subsection{\textbf{Limitations}}  
\subsubsection{\textbf{Computational-Only Validation}}  
Without experimental testing, critical biological aspects such as cellular uptake, metabolic stability, toxicity, immunogenicity, and off-target effects remain unknown. Molecular docking and ADMET predictions provide useful early insights, but their reliance on simplified force fields, static protein models, and approximations of solvation and dynamics may lead to false positives or overestimated binding affinities. Additionally, dependence on AlphaFold3-predicted structures introduces uncertainty, as even high-confidence models may deviate from native conformations, particularly in flexible regions crucial for ligand binding.
\subsubsection{\textbf{Reliance on Bulk RNA-Seq Data}}  
Another limitation stems from using bulk RNA-seq data from the TCGA-LAML cohort. Bulk transcriptomics averages signals across heterogeneous cell populations, potentially masking the molecular profiles of rare leukemic stem cells and therapy-resistant subclones that drive relapses. As a result, target selection may be biased toward dominant clones while overlooking clinically relevant vulnerabilities in minor subpopulations.

\subsection{\textbf{Future Studies}}  
\subsubsection{\textbf{Experimental Validation and Iterative Optimization}} Future work will prioritize experimental validation and iterative optimization of top-ranked ligands. Selected candidates will be synthesized and tested in vitro for binding and potency, followed by in vivo evaluation in patient-derived xenograft models to assess efficacy, pharmacokinetics, pharmacodynamics, and preliminary toxicity. These results will feed into a closed-loop Design-Make-Test-Analyze cycle, enabling continuous refinement of the computational model and medicinal chemistry optimization.
\subsubsection{\textbf{Integration of Single-Cell and Spatial Multi-Omics}} 
To address tumor heterogeneity, future iterations will integrate single-cell RNA sequencing and spatial transcriptomic data. This will allow identification of molecular vulnerabilities in leukemic stem cells and resistant subclones, support subclone-specific targeting and guide the design of combination or multi-target therapies. Incorporating profiles from non-malignant bone marrow cells, such as immune and stromal populations, will enable microenvironment-aware drug design by highlighting targets involved in pro-leukemic interactions and immunomodulatory pathways.

\subsubsection{\textbf{Model Expansion and Generalization}}
The pipeline’s modular, disease-agnostic design allows its extension beyond AML to other hematologic malignancies, solid tumors with high heterogeneity, and diseases with limited treatment options. Integrating additional data modalities, including proteomics, epigenomics, and metabolomics, will provide a more comprehensive representation of disease biology and reveal previously inaccessible druggable targets. Collectively, these features position the pipeline as a flexible, generalizable computational platform for precision drug discovery across oncology and other therapeutic areas with unmet clinical needs.

\section{\textbf{Competing interests}}
All authors declare no financial or non-financial competing interests.

\section{\textbf{Data Availability Statement}}
The TCGA acute myeloid leukemia (LAML) exon expression by RNAseq (polyA+ IlluminaHiSeq) dataset used in this paper is publicly available online at: https://shorturl.at/MnAFM 

\section{\textbf{Code availability}}
The underlying code for this study [and training/validation datasets] is not publicly available but may be made available to qualified researchers on reasonable request from the corresponding author.


\begin{thebibliography}{00}

\bibitem{1}
National Library of Medicine, ``Acute myeloid leukemia,'' MedlinePlus.
[Online]. Available: \url{https://medlineplus.gov/acutemyeloidleukemia.html}

\bibitem{2}
N. M. N. P. Bcop Bcps and A. P. M. Pa-C,
``Acute Myeloid Leukemia: An Ever-Changing Disease,''
\emph{Journal of the Advanced Practitioner in Oncology},
vol. 10, no. 8, Dec. 2019,
doi: 10.6004/jadpro.2019.10.8.12.

\bibitem{3}
M. Morcos-Sandino, S. I. Quezada-Ramírez, and A. Gómez-De León,
``Advances in the Treatment of Acute Myeloid Leukemia: Implications for Low- and Middle-Income Countries,''
\emph{Biomedicines}, vol. 13, no. 5, p. 1221, 2025,
doi: 10.3390/biomedicines13051221.

\bibitem{4}
``Acute myeloid leukemia -- Cancer Stat Facts,'' SEER.
[Online]. Available: \url{https://seer.cancer.gov/statfacts/html/amyl.html}

\bibitem{5}
``What role is precision medicine playing in transforming cancer treatment -- and how are developers adapting?,'' Pharma’s Almanac.
[Online]. Available:
https://www.pharmasalmanac.com/articles/what-role-is-precision-medicine-playing-in-transforming-cancer-\\ treatment-and-how-are-developers-adapting


\bibitem{6}
S. C.-h. Pegg, J. J. Haresco, and I. D. Kuntz,
``A genetic algorithm for structure-based de novo design,''
\emph{Journal of Computer-Aided Molecular Design},
vol. 15, no. 10, pp. 911--933, Oct. 2001,
doi: 10.1023/A:1014389729000.

\bibitem{7}
R. Vasundhara Devi, S. Siva Sathya, and M. S. Coumar,
``Evolutionary algorithms for de novo drug design -- A survey,''
\emph{Applied Soft Computing}, vol. 27, pp. 543--552, 2015,
doi: 10.1016/j.asoc.2014.09.042.

\bibitem{8}
X. Wang, K. Song, L. Li, and L. Chen,
``Structure-Based Drug Design Strategies and Challenges,''
\emph{Current Topics in Medicinal Chemistry},
vol. 18, no. 12, pp. 998--1006, Aug. 2018,
doi: 10.2174/1568026618666180813152921.

\bibitem{9}
S. M. Ajjarapu \emph{et al.},
``Ligand-based drug designing,''
in \emph{Bioinformatics}, D. B. Singh and R. K. Pathak, Eds.
Cambridge, MA, USA: Academic Press, 2022, ch. 15, pp. 233--252,
doi: 10.1016/B978-0-323-89775-4.00018-3.

\bibitem{10}
K. H. Metzeler \emph{et al.},
``Spectrum and prognostic relevance of driver gene mutations in acute myeloid leukemia,''
\emph{Blood}, 2016.
[Online]. Available: \url{https://doi.org/10.1182/blood-2016-01-693879}

\bibitem{11}
B. Samra \emph{et al.},
``Venetoclax-Based Combinations in Acute Myeloid Leukemia: Current Evidence and Future Directions,''
\emph{Frontiers in Oncology}, vol. 10, p. 562558, Nov. 2020,
doi: 10.3389/fonc.2020.562558.

\bibitem{12}
A. Subramanian \emph{et al.},
``A next generation connectivity map: L1000 platform and the first 1,000,000 profiles,''
\emph{Cell}, vol. 171, no. 6, pp. 1437--1452.e17, Nov. 2017,
doi: 10.1016/j.cell.2017.10.049.

\bibitem{13}
Z. Wang \emph{et al.},
``Extraction and analysis of signatures from the Gene Expression Omnibus by the crowd,''
\emph{Nature Communications}, vol. 7, p. 12846, Sep. 2016,
doi: 10.1038/ncomms12846.

\bibitem{14}
``A comprehensive clinically informed map of dependencies in cancer cells and framework for target prioritization,''
\emph{Cancer Cell}, pp. 301--316, 2024.
[Online]. Available: \url{https://doi.org/10.1016/j.ccell.2023.12.016}

\bibitem{15}
R. Gómez-Bombarelli \emph{et al.},
``Automatic chemical design using a data-driven continuous representation of molecules,''
\emph{ACS Central Science}, vol. 4, no. 2, pp. 268--276, Jan. 2018,
doi: 10.1021/acscentsci.7b00572.

\bibitem{16}
D. C. Nicola and T. Kipf,
``MolGAN: An implicit generative model for small molecular graphs,''
arXiv:1805.11973, 2018.
[Online]. Available: \url{https://arxiv.org/abs/1805.11973}

\bibitem{17}
D. A. Erlanson,
``Introduction to Fragment-Based Drug Discovery,''
\emph{Topics in Current Chemistry}, vol. 317, pp. 1--32, Jan. 2011,
doi: 10.1007/128\_2011\_180.

\bibitem{18}
J. H. Jensen,
``A graph-based genetic algorithm and generative model/Monte Carlo tree search for the exploration of chemical space,''
\emph{Chemical Science}, vol. 10, no. 12, pp. 3567--3572, Jan. 2019,
doi: 10.1039/C8SC05372C.

\bibitem{19}
Y. Kwon and J. Lee,
``MolFinder: An evolutionary algorithm for the global optimization of molecular properties,''
\emph{Journal of Cheminformatics}, vol. 13, p. 24, Mar. 2021,
doi: 10.1186/s13321-021-00501-7.

\bibitem{20}
O. Trott and A. J. Olson,
``AutoDock Vina: Improving the speed and accuracy of docking,''
\emph{Journal of Computational Chemistry},
vol. 31, no. 2, pp. 455--461, Jun. 2009,
doi: 10.1002/jcc.21334.

\bibitem{21}
H. Stärk \emph{et al.},
``EquiBIND: Geometric Deep Learning for Drug Binding Structure Prediction,''
arXiv:2202.05146, 2022.
[Online]. Available: \url{https://arxiv.org/abs/2202.05146}

\bibitem{22}
J. Lamb \emph{et al.},
``The Connectivity Map: Using gene-expression signatures to connect small molecules, genes, and disease,''
\emph{Science}, vol. 313, no. 5795, pp. 1929--1935, Sep. 2006,
doi: 10.1126/science.1132939.

\bibitem{23}
``UCSC Xena.''
[Online]. Available: \url{https://xenabrowser.net/datapages/}

\bibitem{24}
``HiSeq 2000 Support,'' Illumina.
[Online]. Available: \url{https://support.illumina.com/sequencing/sequencing_instruments/hiseq_2000.html}

\bibitem{25}
J. Abramson \emph{et al.},
``Accurate structure prediction of biomolecular interactions with AlphaFold 3,''
\emph{Nature}, vol. 630, no. 8016, pp. 493--500, May 2024,
doi: 10.1038/s41586-024-07487-w.

\bibitem{26}
``ProteinsPlus Server -- DOGSite.''
[Online]. Available: \url{https://proteins.plus/help/dogsite}

\bibitem{27}
L. Fu \emph{et al.},
``ADMETlab 3.0: An updated comprehensive online ADMET prediction platform,''
\emph{Nucleic Acids Research}, vol. 52, no. W1, pp. W422--W431, Apr. 2024,
doi: 10.1093/nar/gkae236.

\bibitem{28}
M. Bugnon \emph{et al.},
``SwissDock 2024: Major enhancements for small-molecule docking,''
\emph{Nucleic Acids Research}, vol. 52, no. W1, pp. W324--W332, Jul. 2024,
doi: 10.1093/nar/gkae300.

\bibitem{29}
J. Eberhardt \emph{et al.},
``AutoDock Vina 1.2.0,''
\emph{Journal of Chemical Information and Modeling},
vol. 61, no. 8, pp. 3891--3898, Jul. 2021,
doi: 10.1021/acs.jcim.1c00203.

\bibitem{30}
Y. Liu \emph{et al.},
``CB-Dock2: Improved protein--ligand blind docking,''
\emph{Nucleic Acids Research},
vol. 50, no. W1, pp. W159--W164, May 2022,
doi: 10.1093/nar/gkac394.

\end{thebibliography}
\end{document}